\documentclass{article}

\usepackage{ml4cps}

\usepackage[utf8]{inputenc} 
\usepackage[T1]{fontenc}    
\usepackage{hyperref}       
\usepackage{url}            
\usepackage{soul}
\usepackage{booktabs}       
\usepackage{amsfonts}       
\usepackage{nicefrac}       
\usepackage{microtype}      
\usepackage{makecell}      
\usepackage{amsmath}
\usepackage{xcolor}
\usepackage{cleveref}       
\usepackage{lipsum}         
\usepackage{graphicx}
\usepackage{natbib}
\usepackage{doi}
\usepackage{amsmath}
\usepackage{amssymb}
\usepackage[]{mdframed}
\usepackage[linesnumbered,ruled,vlined]{algorithm2e}
\usepackage{booktabs}
\usepackage{array}
\usepackage{subcaption}

\usepackage[colorinlistoftodos]{todonotes}
\title{Physics-Guided Dual Implicit Neural Representations for 
Source Separation}

\date{}

\usepackage{authblk}
\author[1,2]{Yuan Ni}

\author[1,2,*]{Zhantao Chen}

\author[1,2]{Alexander N.\@ Petsch}

\author[1,3]{Edmund Xu}

\author[2]{Cheng Peng}

\author[4]{Alexander I.\@ Kolesnikov}
\author[5]{Sugata Chowdhury}
\author[6]{Arun Bansil}
\author[1]{Jana B.\@ Thayer}
    
\author[1,2,*]{Joshua J.\@ Turner}

\affil[1]{Linac Coherent Light Source, SLAC National Accelerator Laboratory, Menlo Park, California 94025, USA}
\affil[2]{Stanford Institute for Materials and Energy Sciences, Stanford University and SLAC National Accelerator Laboratory, Menlo Park, California 94025, USA}
\affil[3]{Computer Science and Engineering Department. University of California Santa Cruz, Santa Cruz, CA 95064, USA}
\affil[4]{Neutron Scattering Division, Oak Ridge National Laboratory, Oak Ridge, TN 37831, USA}

\affil[5]{Department of Physics and Astronomy, Howard University, Washington DC, 20059}

\affil[6]{Physics Department, Northeastern University, Boston, MA 02115}

\affil[*]{Correspondence: \href{mailto:zhantao@stanford.edu}{zhantao@stanford.edu} (Z.C.), \href{mailto:joshuat@slac.stanford.edu}{joshuat@slac.stanford.edu} (J.J.T.)}

\newcommand{\ud}{\mathrm{d}}

\hypersetup{
pdftitle={contribution title},
pdfsubject={contribution subject},
pdfauthor={First author, Second author, Third author, Fourth author, Fifth author},
pdfkeywords={First keyword, Second keyword, Third keyword, Fourth keyword, Fifth keyword},
}

\begin{document}

\maketitle

\begin{abstract}
Significant challenges exist in efficient data analysis of most advanced experimental and observational techniques because the collected signals often include unwanted contributions--such as background and signal distortions--that can obscure the physically relevant information of interest.
To address this, we have developed a self-supervised machine-learning approach for source separation using a dual implicit neural representation framework that jointly trains two neural networks: one for approximating distortions of the physical signal of interest and the other for learning the \textit{effective} background contribution. Our method learns directly from the raw data by minimizing a reconstruction-based loss function without requiring labeled data or pre-defined dictionaries. 
We demonstrate the effectiveness of our framework by considering a challenging case study involving large-scale simulated as well as experimental momentum-energy-dependent inelastic neutron scattering data in a four-dimensional parameter space, characterized by heterogeneous background contributions and unknown distortions to the target signal. The method is found to successfully separate physically meaningful signals from a complex or structured background even when the signal characteristics vary across all four dimensions of the parameter space.
An analytical approach that informs the choice of the regularization parameter is presented. Our method offers a versatile framework for addressing source separation problems across diverse domains, ranging from superimposed signals in astronomical measurements to structural features in biomedical image reconstructions.
\end{abstract}


\section{Introduction}\label{Section:1}

Although experimental observations are the cornerstone of physical sciences, the measured data almost always contain unwanted contributions beyond the signals of interest. For example, in astrophysics, astrophysical X-ray and gamma-ray observatories are often overwhelmed by high background levels and noise \cite{Ehlert2022A}. Similarly, X-ray and neutron scattering experiments commonly record signals originating from multiple sources other than the measured sample, such as the scattering from the substrate and other excitations \cite{petsch2023high}. Many existing practices for separating the target signals from the total measured signal are heuristically developed, for instance, selecting a presumed background region to subtract a global background signal, or using simple kernels to perform deconvolution \cite{Piccardi2004,Sarradj2012}. However, real-world experimental data often exhibit greater complexity, including heterogeneous background contributions and unknown distortions to the target signal. 
Consequently, developing a more accurate and effective method to separate the physically meaningful signals from background and noise is critically important for the reliable interpretation of experimental data.
\\


%
\begin{figure}[hbt!]
    \centering
    \includegraphics[width=\linewidth]{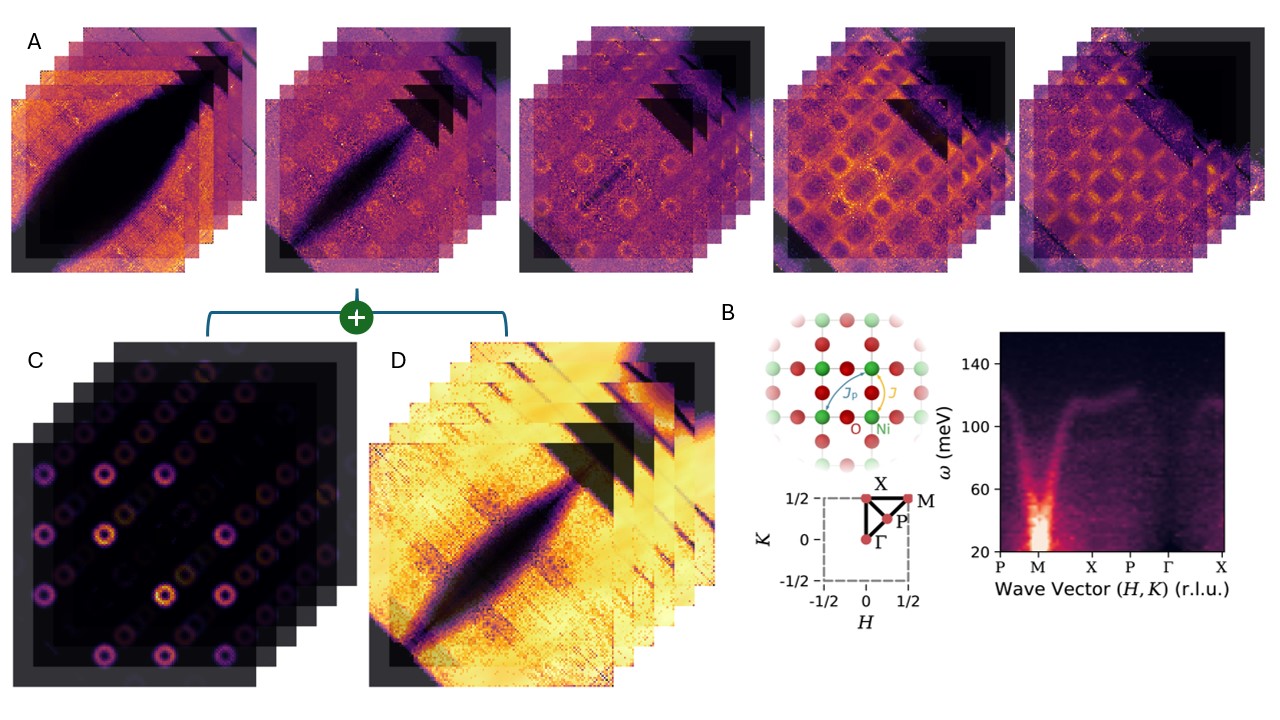}
    \caption{Example of a source separation task in 4D inelastic neutron scattering data, parameterized by $(H, K, L, \omega)$. (A) Raw experimental data across an increasing range of $\omega$ (energy bands) from left to right. Target signal is obscured by high background and noise. (B) Spatial lattice structure of the measured compound, La$_2$NiO$_4$, along with an illustration of the spin model and its parameters used for simulating the dynamic structure factor $\mathbf{S}(\mathbf{Q},\omega)$. Reproduced from \cite{chen2025implicit} under the CC BY-NC license. (C) The separated 4D signal of interest component using the proposed method. Previously hidden features are now revealed clearly for further physical analysis. (D) 4D background component separated from the measured data, capturing the background variations.}
    \label{fig:4D}
\end{figure}

When the measured signal can be modeled by a linear mixture of multiple source signals, the task of signal decomposition generally becomes a special case of source separation, which includes texture extraction, image decomposition, or background/foreground separation. Conventional source separation methods often rely on heuristic approaches \cite{Sarradj2012}, manual pre-processing \cite{bi2019extension}, or strong prior assumptions \cite{bouwmans2015decomposition, liu2022efficient,Candes2011RPCA,Gandelsman2019DoubleDIP}. These might include assumptions about the statistical independence of sources or their sparsity, connectivity, contrast, spatial-temporal and angle-distance variations, and deep-learning based priors, to enhance the accuracy and robustness of separation across different contexts. For instance, Independent Component Analysis (ICA) \cite{COMON1994287} assumes statistical independence between sources. Alternatively, sparsity assumptions are used to improve conventional separation performance by assuming that the data can be represented in a sparse form under certain transformations \cite{YIN2007,Starck2003}.
\\

However, in many scientific applications, these priors impose constraints that are often difficult to satisfy in practice \cite{Lee2011A,Kuruoglu2013dependent}. As a result, conventional source separation methods, such as statistical approaches relying on independence assumptions (e.g., ICA) and sparsity-based methods employing predefined dictionaries, often struggle with scientific data, where sources typically lack statistical independence and are not sparse/low-rank in standard representations. Moreover, data acquired in scientific experiments, such as the data from scattering measurements, may naturally reside in higher dimensional spaces than the image data. For example, the dynamical structure factor $\mathbf{S}(\mathbf{Q}, \omega)$ obtained from inelastic scattering is defined over a four-dimensional domain combining momentum transfer $\mathbf{Q} = (H, K, L)^{\top}$ in reciprocal lattice units (r.l.u.) and energy transfer $\hslash\omega$\footnote{From here onwards, we will use units with $\hslash=1$ for simplicity.}. Other experiments, such as high-energy particle collisions can span $10+$ dimensions. This high dimensionality poses challenges for conventional source separation methods, which are typically designed for images/volumetric data defined on 2D/3D spatial grids. Furthermore, traditional separation methods may not offer analytical solutions and instead rely on iterative numerical solutions, making them computationally expensive and potentially prohibitive when scaling to high-dimensional and large-scale scientific datasets. While specialized source separation methods exist in various scientific fields, including neutron and X-ray spectroscopy \cite{TOUGAARD1989343,VEGH2006159}, digital subtraction angiography in biomedical imaging \cite{crummy2018history}, and astrophysics \cite{Starck2003}, these methods often require domain-specific and physics informed pre-processing and extensive modeling. They are, however, frequently unsuitable for high-throughput, high-dimensional experimental systems that demand more automated and more scalable data processing. 
\\

To address these challenges, we propose a self-supervised machine learning approach for source separation using a dual Implicit Neural Representations (INRs) framework. One representation network learns to transform the simulated data (e.g., the dynamic structure factor, $\mathbf{S}(\mathbf{Q},\omega)$ in INS), or more generally, any spectral function describing the energy--momentum distribution of excitations in a material, into realistic physical signals, implicitly reconstructing a heterogeneous distortion kernel that captures deviations between the measured data and the known model. The other network learns a representation of the variable background and other non-target signal contributions. The two INRs are trained jointly without explicit supervision or labels indicating which parts correspond to signal or background. Both INRs are designed so that each produces a valid representation of the specific signal it is intended for, while being inefficient at representing the other components. Thus, when combined into a final representation, the sum of these two individual INRs is expected to yield a reconstruction with a legitimate separation of sources. Moreover, when a known physical model is available for any component of the source signals, we can incorporate it into the architecture of the corresponding INR, enabling a hybrid modeling approach where the network learns only the unmodeled part. Unlike conventional source separation techniques, our approach does not require labeled data or predefined transformations for signal source separations. Instead, it learns to disentangle signal sources from the raw data itself by minimizing a regularized reconstruction loss.\\


We demonstrate the efficacy of our method by considering a use case where a variety of different source signals are present in the raw measurements, which involve the 4D parameter space of inelastic neutron scattering (INS) experiments from the square-lattice spin-1 antiferromagnet, La\textsubscript{2}NiO\textsubscript{4}. Our goal is to extract the single-magnon scattering signal from the measured signal which also contains various contributions from multi-magnon, phonon, and elastic scattering processes. {Simulations of the single-magnon distribution are treated by the network to differentiate between these other signals, which could be the signal of interest for other types of scientific objectives.}
A conventional data analysis pipeline often involves estimating the background using a common subtraction prior, wherein a region of the experimental data, assumed to contain purely non-physically relevant signals, is used as a reference to estimate and subtract the global background.
In contrast, we show that our method can more effectively handle heterogeneous backgrounds and model unknown distortions by taking full advantage of simulated INS spectroscopy data using an INR-based convolutional kernel, capturing 4D momentum-energy coordinate-dependent distortions, while learning all sample-related signals through an additional background INR.\\

The remainder of this paper is organized as follows: Section \ref{Section:2} presents the proposed dual INR framework and describes how it incorporates forward modeling. Section \ref{Section:2} also discusses the inductive bias of INRs for representing distinct signal sources. In Section \ref{Section:3}, we demonstrate numerically that the proposed framework achieves successful separation in both the simulated and experimental INS data. Section \ref{Section:3} also discusses physics-guided hyperparameter selection and regularization. Section \ref{Sec:4} discusses the effectiveness of our approach in noise removal, and as a data compression method as a side benefit. In this section, we also outline potential future directions.

\section{Problem Formulations \label{Section:2}}

\subsection{Forward Model}

Many experimental measurements contain contributions from multiple sources. Consider the case where the signals are a linear combination of two sources and are corrupted by Poisson noise,
\begin{equation}\label{eq:0}
\mathbf{S}_\text{expt}^{\ast} \propto  \mathrm{Poisson}(\bar{\mathbf{S}}_{\text{expt}}^{\ast}),\quad \bar{\mathbf{S}}_{\text{expt}}^{\ast}=\mathbf{S}^{(1)}_{\text{sig}} + \mathbf{S}^{(2)}_{\text{sig}},
\end{equation}
where $\mathbf{S}^{(1)}_{\text{sig}}$ and $\mathbf{S}^{(2)}_{\text{sig}}$ correspond to ideally decoupled signals from different sources, and $\bar{\mathbf{S}}^{\ast}_{\text{expt}}$ denotes the underlying noiseless total signal. This model captures many realistic experimental setups where one source represents the physically meaningful signal from target excitation or target sample, and the second signal encompasses all other contributions. The goal is to accurately and efficiently extract these two individual parts from the overall measurement to gain insights into each source.
\\

Since the only observable is $\mathbf{S}^{\ast}_{\text{expt}}$, $\mathbf{S}^{(1)}_{\text{sig}}$ and $\mathbf{S}^{(2)}_{\text{sig}}$ can, in principle, have arbitrary relative intensities. Such an inverse problem is impossible to solve without further assumptions. For example, if $(\mathbf{S}^{(1)}_{\text{sig}},\mathbf{S}^{(2)}_{\text{sig}})$ is a pair of solutions, then for any arbitrary component $\mathbf{S}^{(0)}$,  $(\mathbf{S}^{(1)}_{\text{sig}} + \mathbf{S}^{(0)},\mathbf{S}^{(2)}_{\text{sig}} -\mathbf{S}^{(0)})$ also gives a solution. A basic assumption that enables a meaningful separation is that the constituent signals $(\mathbf{S}^{(1)}_{\text{sig}},\mathbf{S}^{(2)}_{\text{sig}})$ originate from incoherent sources that can be represented by distinct representations (e.g. predefined dictionaries in sparse coding or learned INRs). In other words, each representation should efficiently capture the specific signal it is designed for, while suppressing representations of the other components.  However, since the separation task in \eqref{eq:0} is fundamentally underdetermined, we must leverage the fact that such data possess low intrinsic dimensionality. 
\\

When the source signal can be at least partially modeled, one can incorporate such prior knowledge to guide signal separation. A common scenario in imaging systems involves modeling the measured signal
\begin{equation}\label{eq:general}
\mathbf{S}_{\text{expt}}^{\ast}=\mathbf{S}^{(1)} * \boldsymbol{\kappa} + \mathbf{S}^{(2)}_{\text{sig}} + \mathbf{N},
\end{equation}
where $\boldsymbol{\kappa}$ is a convolution kernel representing the system response or distortion (e.g., point spread function (PSF) of an imaging system), and $\mathbf{N}$ denotes the noise, modeled using the normal approximation to a Poisson distribution. As a concrete example, in astrophysics, $\mathbf{S}^{(1)}$ may correspond to discrete sources such as stars, while $\mathbf{S}^{(2)}_{\text{sig}}$ represents the diffuse sky background. Another example arises in digital subtraction angiography (DSA), where $\mathbf{S}^{(1)}$ can represent the vascular structures of interest, and $\mathbf{S}^{(2)}_{\text{sig}}$ represents background tissues and motion artifacts. 

\subsection{Dual-INR Representation: Modeling Each Component with a Separate Neural Network}
In our demonstration example of INS, the measured signal consists of a physically meaningful single-magnon-related component, while the background includes many other signals, such as phonons that are not of immediate interest for this experiment, in addition to the noise component. The physically meaningful part is subject to resolution broadening due to the finite energy resolution of the instrument. This broadening can be modeled as the convolution between the sharp theoretical peak and the instrumental energy resolution function (i.e. $\boldsymbol{\kappa}$), typically assumed to be Gaussian. However, in real-world experiments, the peak shape might suffer from unknown distortions and even be asymmetric, beyond the expression capability of simple energy-broadening kernels. 
\\

To address this limitation, we model the distorted physical signal as a convolution of the ideal signal with a spatially varying, learnable kernel $\boldsymbol{\kappa}$. In the specific example of the INS of a square-lattice spin-1 antiferromagnet, we can simulate the ideal single-magnon INS spectrum using the linear spin wave theory, i.e., $\mathbf{S}_{\text{sim}}(\mathbf{Q},\omega; J, J_p)$ as a function of momentum transfer $\mathbf{Q}$, energy transfer $\omega$ and fixed parameters $J$ and $J_p$. Generalizing the idea of convolution with a Gaussian kernel, we model the distorted physical signal as a convolution between the ideal single-magnon signal and a spatially-varying and learnable kernel, namely, 
\begin{equation}
    \mathbf{S}^{(1)}_{\text{sig}}(\mathbf{Q},\omega)=\int \mathbf{S}_{\text{sim}}(\mathbf{Q}^{\prime},\omega^{\prime};J,J_p)\, \boldsymbol{\kappa}_{r,\hat{\phi}}(\mathbf{Q}^{\prime}-\mathbf{Q},\omega^{\prime}-\omega, \mathbf{Q}, \omega) \ud\mathbf{Q}^{\prime}\,\ud\omega^{\prime},
\end{equation}
where $\boldsymbol{\kappa}_{r,\hat{\phi}}$ represents a learnable, $(\mathbf{Q},\omega)$-dependent kernel within a neighborhood of size $\sim r^{d}$, where $r$ is a user-defined hyperparameter controlling the window size of the kernel and $d$ is the dimension of the space that data lives in; for INS this data, $S(\mathbf{Q},\omega)$, we have $d=4$.
\\

We propose to use two INRs to represent the two distinct parts of the signal, namely the $\boldsymbol{\kappa}$ and $\mathbf{S}^{(2)}_{\text{sig}}$. The basic idea underlying the INR is to use a neural network to learn representations of the signal. INRs are neural networks (often MLPs) that represent data as continuous functions that are trained to parameterize signals by taking coordinates as inputs, applying a feature mapping, and predicting the associated data values at each coordinate,
\begin{equation} 
    f_{\hat{\theta}}(\mathbf{x}) \approx \mathbf{S}(\mathbf{x}), 
\end{equation}
where $\mathbf{x} \in\mathbb{R}^D$ denotes a D-dimensional coordinate (e.g., a 2D pixel location in a digital image or high-dimensional coordinate in experimental data), $\mathbf{S}(\mathbf{x})$ is the signal value at that coordinate (such as the pixel intensity or volumetric data), and $f_{\hat{\theta}}$ is a neural network with parameters ${\hat{\theta}}$ trained to approximate the signal. $f_{\hat{\theta}}$ encodes the data in its network parameters, which can yield a much more compact description. This idea leverages the fact that many real-world signals are highly structured and thus compressible \cite{essakine2024we,essakine2024we}.
\\

In this work, the kernel values are represented by an INR (i.e., $\boldsymbol{\kappa}_{r,\hat{\phi}}: (\mathbf{Q}^{\prime}-\mathbf{Q},\omega^{\prime}-\omega, \mathbf{Q}, \omega) \mapsto \mathbb{R}^{N_{\boldsymbol{\kappa}}}$ with $N_{\boldsymbol{\kappa}}\sim r^{d}$), which is parameterized by a set of trainable weights $\hat{\phi}$. We represent $\mathbf{S}_{\text{sim}}$ using a pre-trained neural network (NN) which maps the combined momentum-energy-parameter input $(H, K, L, \omega, J, J_p)$ to the corresponding value $\mathbf{S}_{\text{sim}}(H, K, L, \omega; J, J_p)$. The network is adopted from Reference~\citenum{chen2025implicit}, which provides further details on its architecture and training.
\\

Following the ideas presented in previous works \cite{chen2025implicit, chitturi2023capturing}, we train an INR to implicitly represent the smooth component $\mathbf{S}^{(2)}_\text{sig}$. That is, we assume that $\mathbf{S}^{(2)}_\text{sig} \approx \mathbf{B}_{\hat{\theta}}$, where $\mathbf{B}_{\hat{\theta}}$ denotes a neural network parameterized by ${\hat{\theta}}$. Thus, even when the exact form of the unknown component is unavailable, modeling it with a compact neural network imposes a useful prior favoring smooth and structured patterns over random noise \cite{Berg_Nyström_2021}.
\textcolor{blue}{}
\\

Therefore, the total signal can finally be approximated by:
\begin{equation}
\bar{\mathbf{S}}_{\text{expt}}^{\text{pred}}(\mathbf{Q},\omega) = \mathbf{B}_{\hat{\theta}}(\mathbf{Q},\omega) + \int \mathbf{S}_{\text{sim}}(\mathbf{Q}^{\prime},\omega^{\prime};J,J_p) \boldsymbol{\kappa}_{r,\hat{\phi}}(\mathbf{Q}^{\prime}-\mathbf{Q},\omega^{\prime}-\omega,\mathbf{Q},\omega)\ud\mathbf{Q}^{\prime}\ud\omega^{\prime}.
\end{equation}

To summarize, we have defined $\mathbf{S}_{\text{sim}}$ as the estimated or simulated ideal signals of interest, while $\boldsymbol{\kappa}_{r, \hat{\phi}}$ and $\mathbf{B}_{\hat{\theta}}$ are adaptively learned from the raw measurements themselves. The reconstructed signal component is given by the convolution $\hat{\mathbf{S}}^{(1)}_{\text{sig}} = \mathbf{S}_{\text{sim}} * \boldsymbol{\kappa}_{\hat{\phi}}$, and the background component is given by $\hat{\mathbf{S}}^{(2)}_{\text{sig}} = \mathbf{B}_{\hat{\theta}}$. Additional details regarding the neural network architecture and training can be found in ~\nameref{Appendix B}.
\\

Rather than employing a basic transformation kernel such as a Gaussian kernel for $\boldsymbol{\kappa}_{\hat{\phi}}$, this learnable kernel setup can model the distortion or non-isotropic spreading of the peaks more effectively. 
This convolved signal part introduces a structured prior on the source by directly incorporating the underlying physical model. A similar approach of representing different parts of a signal using dual INRs can be found in \cite{Roddenberry2023}, where the authors propose the decomposition of a 1D signal into a smooth part and an auxiliary component.
\\

Finally, and most critically, to achieve image source separation in a self-supervised manner, we minimize the following loss function:
\begin{equation}
\mathcal{L} = \| \mathbf{S}_{\text{expt}}^{\ast} - \bar{\mathbf{S}}_{\text{expt}}^{\text{pred}} \|_2 + \lambda \|\mathbf{B}_{\hat{\theta}}\|_2.\label{eq:loss_fct}
\end{equation}

This introduces the regularization parameter $\lambda$ that balances the data fidelity term and the total intensity in the background net. In particular, it encourages the model to represent the single-magnon signal primarily through the convolutional pathway involving the learnable kernel $\boldsymbol{\kappa}_{r,\hat{\phi}}$ by penalizing the prediction magnitudes of $\mathbf{B}_{\hat{\theta}}$. Although the single-magnon component is expected to be captured by the convolutional model, the $\mathbf{B}_{\hat{\theta}}$ can still inadvertently learn single-magnon-related information, leading to unwanted mixtures due to the ill-posed nature of the problem. The regularization term helps prevent such leakage by discouraging $\mathbf{B}_{\hat{\theta}}$ from fitting components that should be attributed to the structured signal. By optimizing this loss function, the proposed method learns to decompose the signal into two parts: namely, the single-magnon-originated signal and all other collected information. 
Because of the implicit bias of INRs and the incorporation of physics modeling, the objective naturally suppress noise as a by-product \cite{Berg_Nyström_2021,essakine2024we}.

\subsection{Optimal selection of the regularization parameter}\label{sec:ambiguity}

The optimal choice of $\lambda$ should balance the data fidelity term and the regularization term. One natural choice is to approximate $
\lambda \approx \frac{\|\mathbf{N}\|_2}{\|\mathbf{S}^{(2)}_{\text{sig}}\|_2}
$, which suggests $\lambda$ scales with the ratio of noise magnitude to the magnitude of the second signal component. In practice, this exact ratio is not known in advance and would typically be determined through cross-validation.
\\

If we have additional knowledge about either one of the sources, we can get a reasonable estimate of the regularization parameter $\lambda$. For example, suppose the signal of interest $\mathbf{S}^{(1)}_{\text{sig}}$ is known to have finite support $\Omega$, as is common in spectroscopy (localized spectral peaks) or astrophysics (bright point sources like stars). Additionally, the background $\mathbf{S}^{(2)}_{\text{sig}}$ is often assumed to be relatively smooth or slowly varying and the noise characteristics are approximately uniform across the field of view. Under these assumptions, we can obtain a reasonable estimate of $\lambda$ using $\lambda \approx \frac{\|\mathbf{N}_{\Omega^c}\|_2}{\|\mathbf{S}^{(2)}_{\text{sig}}|_{\Omega^c}\|_2}$, which relies on information outside the support of $\mathbf{S}^{(1)}_{\text{sig}}$, where only the background and the noise contribute to the signal. The noise level can be implicitly estimated from fitting a single INR to the entire observed data, where the reconstruction residual approximates the noise component due to the implicit low-frequency bias of INRs, i.e. $\left\|\mathbf{N}_{\Omega^c}\right\|_2 \approx \left\|(\mathbf{S}^{\ast}_{\text{expt}} - f_{\text{approx}}(\mathbf{x}))\big|_{\Omega ^ c} \right\|_2$.  Furthermore, since the background is assumed to be smooth with only low-frequency contributions, we estimate $\mathbf{S}^{(2)}_{\text{sig}}\vert _{\Omega^c}$ by applying a low-pass filter as a practical way to estimate the total background component. Hence, we estimate 
\begin{equation}\label{lambda}
    \lambda^* \approx \frac{\left\|(\mathbf{S}^{\ast}_{\text{expt}} - f_{\text{approx}}(\mathbf{x}))\big|_{\Omega^c}\right\|_{2}}{\left\|\text{LPF}(\mathbf{S}^{*}_{\text{expt}}) \big|_{\Omega^c} \right\|_{2}}.
\end{equation}

Since $\mathbf{S}^{(1)}$ vanishes in $\Omega^c$, we have
$\mathbf{S}^*_{\text{expt}}|_{\Omega^c} = \mathbf{S}^{(2)}_{\text{sig}}|_{\Omega^c} + \mathbf{N}_{\Omega^c}$. The choice of $\lambda^*$ in \eqref{lambda} enforces a uniform energy distribution in the recovered background component $\hat{\mathbf{S}}^{(2)}_{\text{sig}}$ across both the support and its complement, i.e. 
$ \frac{\left\|\hat{\mathbf{S}}^{(2)}_{\text{sig}} \big|_{\Omega}\right\|_2}{\left\|\hat{\mathbf{S}}^{(2)}_{\text{sig}} \big|_{\Omega^c}\right\|_2} \approx \frac{\left\|\mathbf{N}_\Omega \right\|_2}{\left\|\mathbf{N}_{\Omega^c}\right\|_2}. $

\section{Results of Source Separation in Inelastic Neutron Scattering Data \label{Section:3}}
\subsection{Challenges of decouple signals in inelastic neutron scattering data} 

We now detail the specific challenges inherent in the experimental x-ray and neutron data that our approach aims to address. A primary difficulty lies in accurately distinguishing faint signals—such as single-magnon, multi-magnon, and coherent or incoherent phonon excitations—from the pervasive and often dominant background. This background typically includes contributions from the elastic line due to diffuse scattering and incoherent scattering from disorder, which can obscure subtle spectral features. The challenge is magnified when the signal-to-noise ratio (SNR) is low, and the signals themselves lack periodicity across all measured axes, thereby limiting the effectiveness of conventional filtering techniques such as Fourier-based high-pass filtering.

Furthermore, in many experimental scenarios, samples are strongly coupled to their environments or inherently produce large backgrounds due to phenomena like intense Bragg peaks near weak charge ordering, diffuse scattering, or other incoherent scattering processes, making it impractical to perform clean background subtraction through reference measurements alone. Additionally, spectral peaks in dispersive systems rarely conform strictly to Voigt profiles, as they result from convolutions of Gaussian contributions (due to instrumental resolution and mosaicity) and Lorentzian contributions (reflecting finite lifetimes). This convolution necessitates accurate kernel deconvolution, which traditional fitting methods often struggle to achieve. Consequently, our approach is validated using both simulated data and experimental datasets in which these multifaceted separation tasks are particularly challenging and existing methods prove inadequate, or at least sub-optimal in most cases.

\subsection{Synthetic data}
We first evaluate our method on synthetic data generated using simulated signals $\mathbf{S}_{\text{sim}}$ with parameters $J=32.0\,\text{meV}$ and $J_p=-2.6\,\text{meV}$ for the magnetic system presented in Ref.\@ \citenum{petsch2023high}. Here, we chose a window size $r =3 $, which corresponds to local 4D kernels of support size $(2r+1)^{4}=7 \times  7 \times 7 \times 7$ for generating the learnable convolution operation. We generated the background component using the reconstructed background obtained from training on the experimental raw data. 
The synthetic data is then constructed as:
\begin{equation}
    \mathbf{S}_{\text{syn}}(\mathbf{Q},\omega) =\mathbf{B}_{\hat{\theta}}(\mathbf{Q},\omega) + \int_{\Omega} \mathbf{S}_{\text{sim}}(\mathbf{Q},\omega;J,J_p) \boldsymbol{\kappa}_{r,\hat{\phi}}(\mathbf{Q}^{\prime}-\mathbf{Q},\omega^{\prime}-\omega,\mathbf{Q},\omega)\,\ud\mathbf{Q}^{\prime}\,\ud \omega^{\prime}.
\end{equation}
Since both the background component $\mathbf{B}_{\hat{\theta}}$ and the physical signal $\mathbf{S}_{\text{sim}}$ used to create the synthetic data are known, we can directly quantify the reconstruction accuracy of our method for each component. 
We estimate the optimal regularization parameter $\lambda^*$ using Equation~\eqref{lambda}, where the LPF is chosen to be a Gaussian filter (standard deviation 5), which yields an analytically estimated value of $\lambda^* = 0.005$.\\

Next, we conducted a comprehensive benchmarking study across different combinations of the two hyperparameter, $\{r,\lambda\}$. The results are summarized in Table \ref{tab:hist_synthetic}. The configuration $\{r = 2, \lambda = 0.0005\}$ is seen to yield the best reconstruction loss, while $\{r = 3, \lambda = 0.005\}$ achieves the most accurate recovery of the signal component (smallest signal deviation in MSLE-sig). Notably, the latter setting matches both the true kernel window size used in the synthetic data generation as well as the analytically derived $\lambda^*$. Larger kernels can model complex signal distortions more accurately, but may also be prone to overfitting to unwanted signal contributions, potentially degrading overall reconstruction results.\\

\begin{figure}[hbt!]
    \centering
    \includegraphics[width=\linewidth]{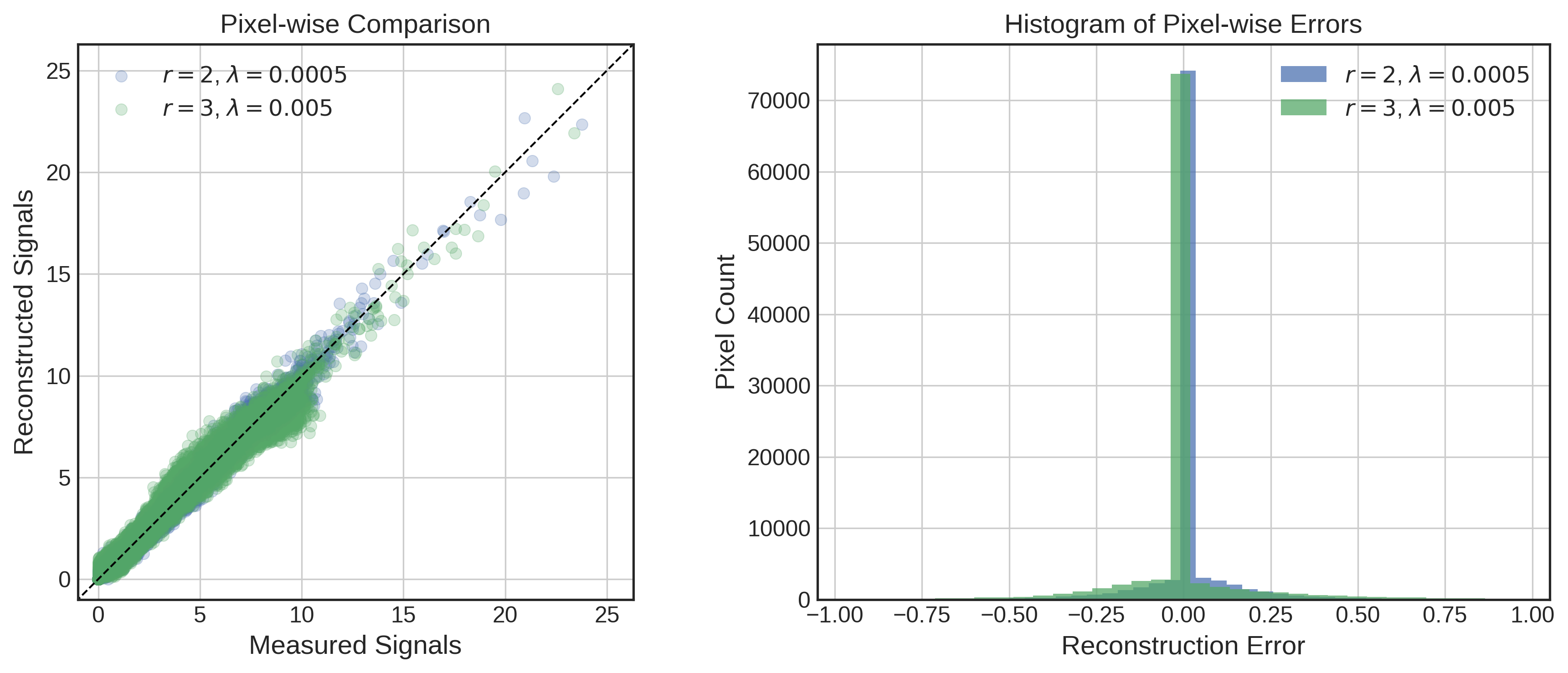}
    \caption{Scatter and histogram plots of flattened pixel-wise differences between the reconstructed and raw data for the synthetic data, using configurations $(r = 2, \lambda = 0.0005)$ and $(r = 3, \lambda = 0.005)$.}
    \label{fig:hist_synthetic}
\end{figure}

We visualize the reconstruction results for the two configurations that yield the best reconstructions in Figure \ref{fig:hist_synthetic}. The tight clustering of points along the line $y = x$ in the 2D histograms in the left panel, which compares flattened pixel-wise true and reconstructed signal intensities, highlights the accuracy and reliability of our method by indicating near-perfect reconstruction. The histogram shows that pixel-wise reconstruction differences are predominantly within ±0.25 intensity units, demonstrating high reconstruction accuracy.

\begin{table}[hbt!]
    \centering
\begin{tabular}{rlllllll}
\toprule
\makecell{$r$} & $\lambda$ & RMSE\textsuperscript{1} & MSLE-sig & MAE-bkg & PSNR & SSIM & MAE  \\
\midrule
2 & 0.0005 & 0.140 & 0.003 & 0.056 & 49.612 & 0.996 & 0.047 \\
2 & 0.005 & 0.162 & 0.002 & 0.060 & 48.292 & 0.996 & 0.054 \\
2 & 0.05 & 0.206 & 0.003 & 0.084 & 46.238 & 0.994 & 0.058 \\
3 & 0.005 & 0.218 & 0.001 & 0.082 & 45.729 & 0.993 & 0.074 \\
3 & 0.0005 & 0.406 & 0.002 & 0.143 & 40.329 & 0.975 & 0.148 \\
3 & 0.05 & 0.467 & 0.012 & 0.184 & 39.123 & 0.971 & 0.115 \\
4 & 0.005 & 0.547 & 0.002 & 0.201 & 37.744 & 0.958 & 0.197 \\
4 & 0.05 & 0.579 & 0.044 & 0.223 & 37.249 & 0.958 & 0.148 \\
4 & 0.0005 & 0.737 & 0.002 & 0.224 & 35.155 & 0.965 & 0.137 \\
\bottomrule
\end{tabular}
    \caption{Summary of reconstruction statistics for using different kernel window sizes $r$ and regularization weights $\lambda$ for the synthetic data experiment. RMSE, PSNR, SSIM, and MAE are computed over the entire reconstructed 4D signal. MSLE-sig measures the mean squared log error between the extracted 4D convoluted signal component and the ground-truth synthesized 4D signal of interest. MAE-bkg reports the mean absolute error between the extracted 4D background and the known simulated 4D background.}
    \label{tab:hist_synthetic}
\footnotesize{\textsuperscript{1}RMSE: Root Mean Squared Error; PSNR: Peak Signal-to-Noise Ratio; SSIM: Structural Similarity Index; MAE: Mean Absolute Error.}
\end{table}

\subsection{Experimental data}
For a detailed account of the data collection and processing methods used in our specific experimental technique, we refer the reader to Refs.\@ \cite{petsch2023high,chitturi2023capturing,chen2025implicit}. The datasets shown have been reconstructed from experimental spectra folded along the diagonal axis within the $(H,K)$-plane. 
This folding approach was used due to the inherent symmetry of our detector setup with respect to the analyzed momentum range. Although folding may result in some loss of direct information along the third momentum axis, $(0,0,L)$, this information can be effectively reconstructed from the available momentum space by considering different sample angles and orientations, as elaborated elsewhere \cite{chen2025implicit}. Understanding this reconstruction process is essential for correctly interpreting momentum-dependent features observed in the presented data.

\begin{figure}[hbt!]
    \centering

    \begin{subfigure}[a]{\linewidth}
        \includegraphics[width=\linewidth]{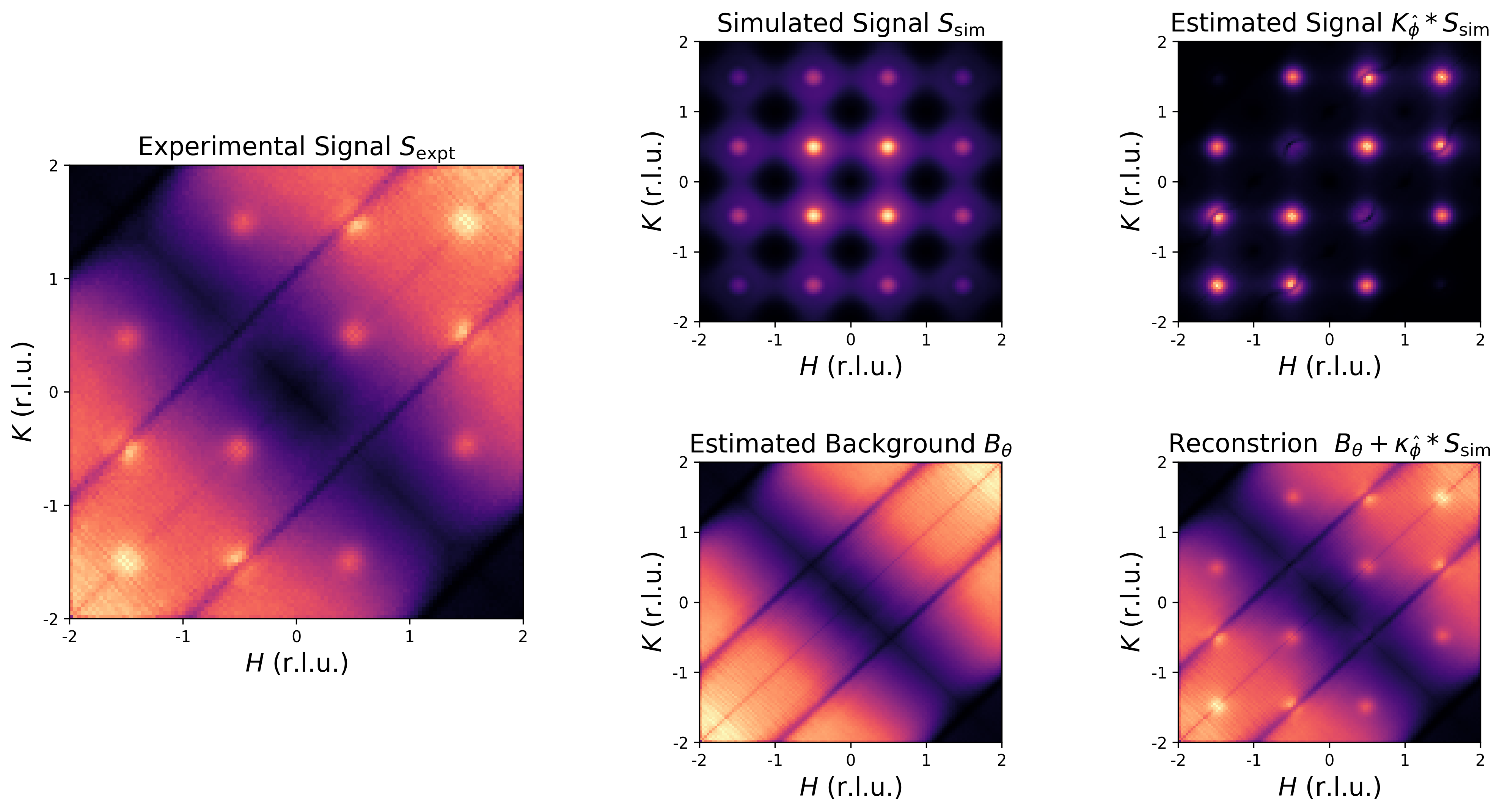}
       \caption{}
        \label{fig:subfig-a}
    \end{subfigure}

    \vspace{0.5em} 

    \begin{subfigure}[b]{\linewidth}
        \includegraphics[width=\linewidth]{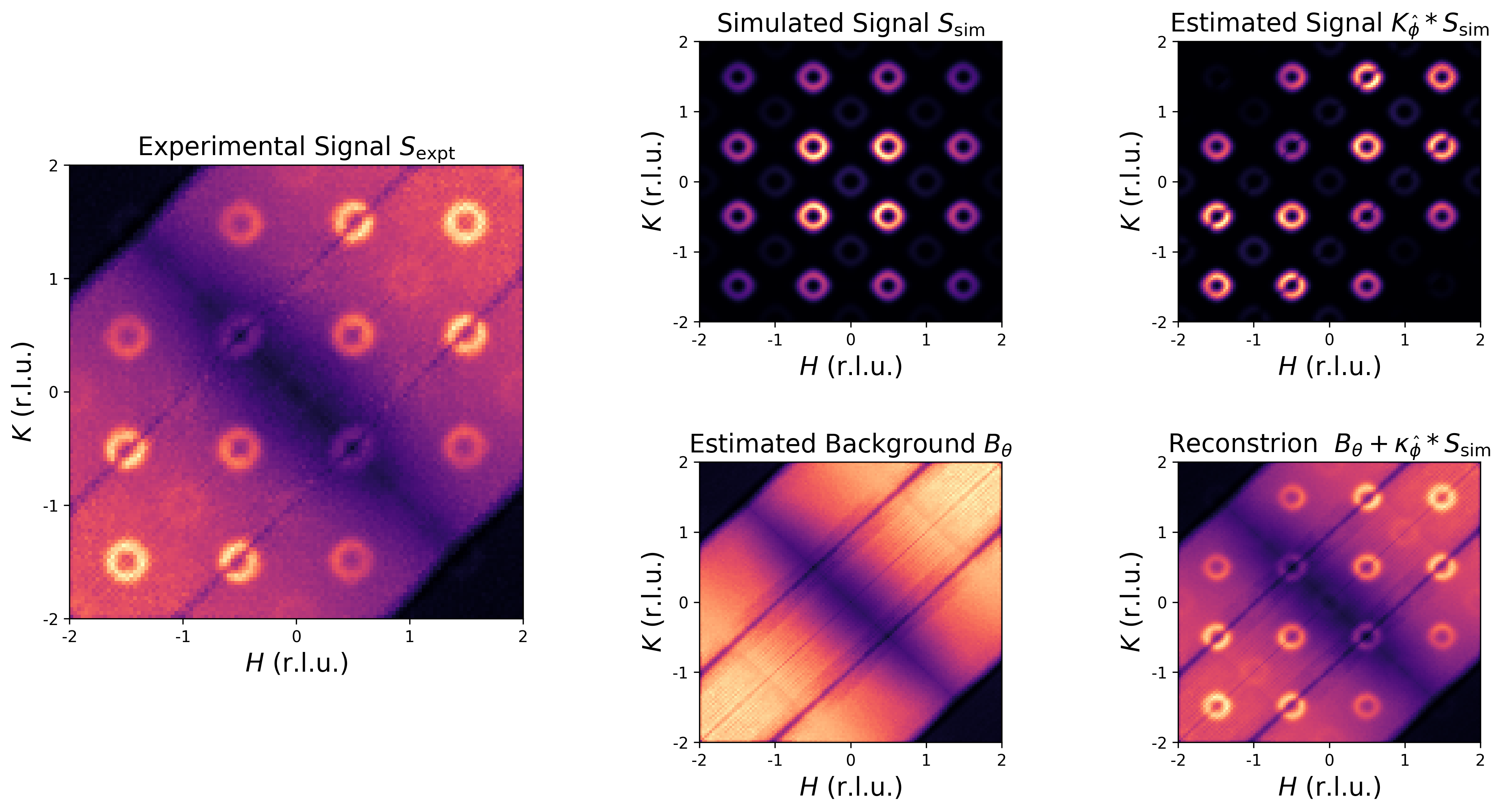}
        \label{fig:subfig-b}
        \caption{}
    \end{subfigure}

    \caption{\textbf{Example of image source separation for an INS dataset in a representative $(H,K)$-slice from the La\textsubscript{2}NiO\textsubscript{4} experiment.} Plots show the best reconstruction using the optimal parameters ($r = 2$, $\lambda = 0.0005$), with (a) all components summed over $l$ and energy $\omega$. Background subtraction used $r=2$, and $\lambda=5 \times 10^{-4}$.
    Shown are: the extracted signal $\sum_{L,\omega} \hat{\mathbf{S}}^{(1)}_{\text{sig}}(H,K,L,\omega)$, background $\sum_{L,\omega} \hat{\mathbf{S}}^{(1)}_{\text{sig}}(H,K,L,\omega)$, their sum $\sum_{L,\omega} (\hat{\mathbf{S}}^{(1)}_{\text{sig}} + \hat{\mathbf{S}}^{(2)}_{\text{sig}})$, the simulated signal $\sum_{L,\omega} \mathbf{S}_{\text{sim}}$, and the measured data $\sum_{L,\omega} \mathbf{S}^*_{\text{expt}}$. This example highlights successful separation of two signal components of distinct physical origin. (b) The exact same presentation as shown in (a), but take instead of integrating over all frequencies where the elastic line is dominating the signal, the energy window here is given by the range of $\omega \in $[60,90]\,meV. }
    \label{fig:composite-separation}
\end{figure}

The final results obtained by the application of our method applied to the 4D INS dataset for source separation are presented in Figure~\ref{fig:composite-separation} and demonstrate the utility of our general framework on a particularly challenging dataset. In Figure~\ref{fig:composite-separation}a, the raw experimental data—comprising contributions from multiple sources—is shown alongside the separation outcomes in which the physically meaningful signal represented by the estimated convolved signal has been effectively isolated from the estimated background. The estimated convolved signal, which is important for subsequent physics analyses, is displayed in the top-right panel. Figure~\ref{fig:composite-separation}b depicts the same data except that it is integrated over a smaller energy window of $\omega \in [60, 90]$\,meV. As the elastic line dominates in the full-range data in panel (a), the reduced bandwidth in panel (b) reveals the signal peaks more distinctly as annular features. 
This captures the main results of this work and will be discussed further below.

\subsection{Connecting Theory and Observation}
The following section provides a detailed analysis of hyperparameter sensitivity, specifically examining how $\lambda$ and $r$ influence the separation results.

\subsubsection{Choice of $\lambda$}
\begin{figure}[hbt!]
    \centering
    \includegraphics[width=\linewidth]{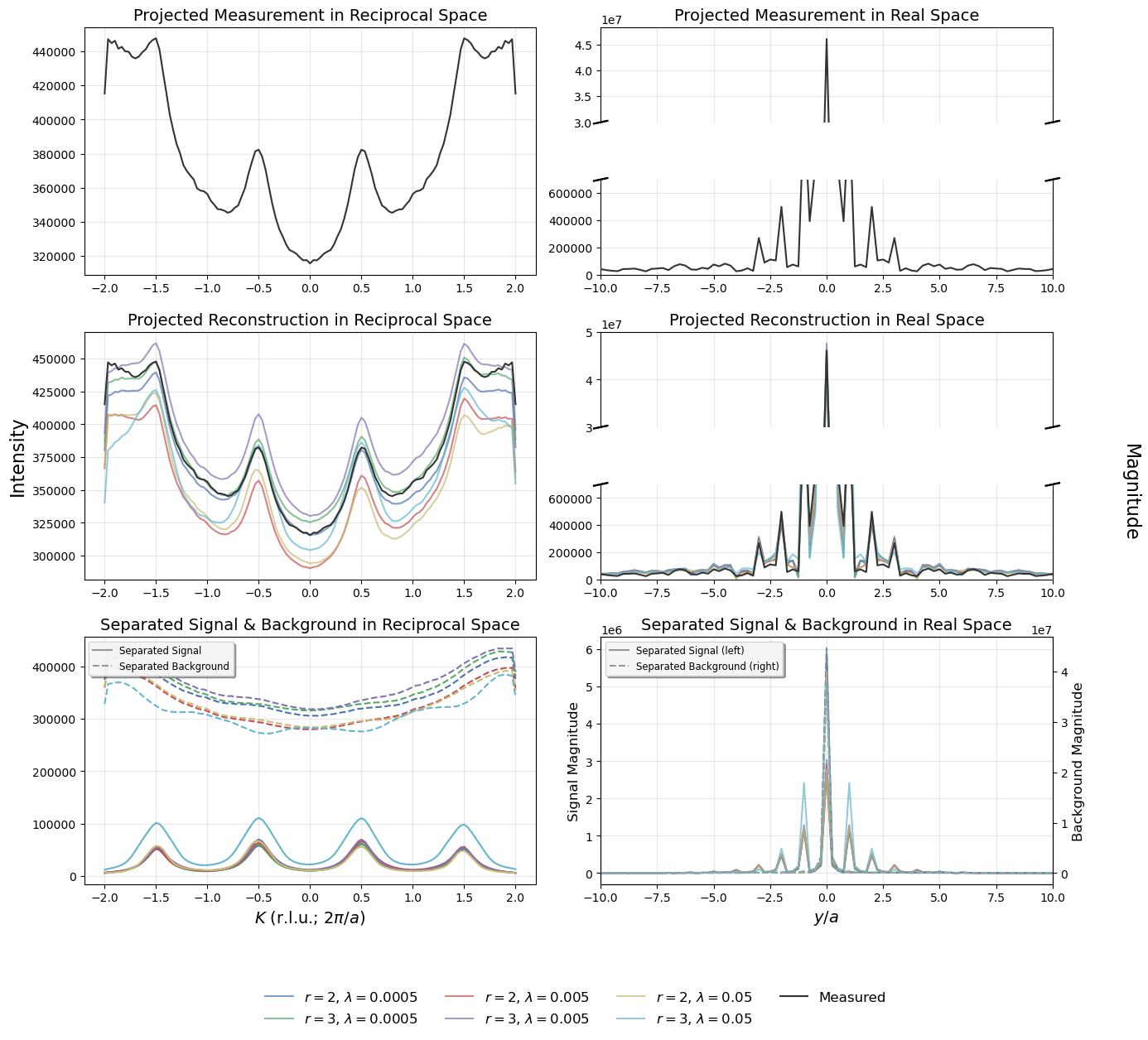}
    \caption{\textbf{Decomposition analysis with varying regularization parameters and support of the kernel net.} The first row shows the Radon transform (line-integral) of the 2D raw measurement $\mathcal{R}_{\alpha = 0}[\sum\limits_{L}\sum\limits_{w}\mathbf{S}^*_{\text{expt}}(H,K,L,\omega)]$ and its Fourier transformation magnitudes. According to the Central Slice Theorem, this represents a slice of the 2D Fourier transformation of $\mathcal{F}[\sum_{L,\omega}\mathbf{S}^*_{\text{expt}}(H,K,L,\omega)](\xi_H=0)$. The left column shows the intensity of the projections while the right column shows the corresponding Fourier transform slices magnitudes. The top row are the projections and slices from the raw measurements, the middle row shows the projections and slices of the reconstructed signals (i.e., $\hat{\mathbf{S}}^{(1)}_{\text{sig}} + \hat{\mathbf{S}}^{(2)}_{\text{sig}}$),  and the bottom row shows the separated components (i.e., $\hat{\mathbf{S}}^{(1)}_{\text{sig}}$ and $\hat{\mathbf{S}}^{(2)}_{\text{sig}}$). Different combinations of the $\lambda$ and $r$ are used.}
    \label{fig:fourier_all}
\end{figure}

\begin{figure}[hbt!]
    \centering
    \includegraphics[width=\linewidth]{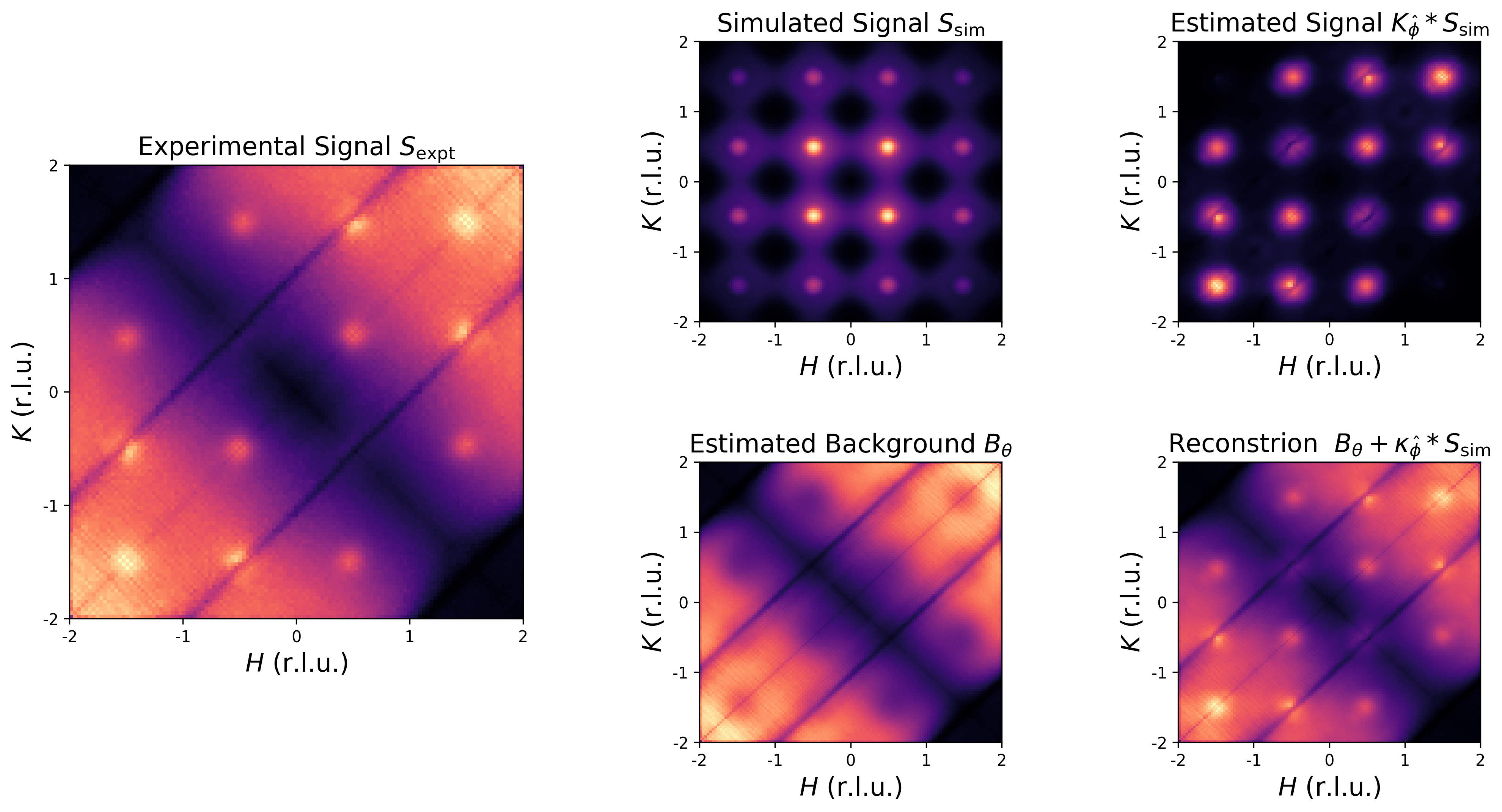}
    \caption{\textbf{Reconstruction plots using $r=3$ and $\lambda = 0.05$.} Visual leakage of the intensity from the background part to the signal part results in low-signal blobs and non-smooth artifacts in the background component. Also, too much intensity is presented in the signal part which can be observed by two overly bright peaks around the center of $\mathbf{S}_{\mathrm{sim}}$.}
    \label{fig:leakage}
\end{figure}

The value of $\lambda$ controls the energy distribution between the two parts of the signal. A larger $\lambda$ penalizes large values of $\left\|\mathbf{B}_{\hat \theta}\right\|_2$, resulting in a low energy reconstruction in the background component $\left\|\hat{\mathbf{S}}^{(2)}_{\text{sig}}\right\|_2$. As discussed in Section \ref{sec:ambiguity}, the optimal $\lambda$ should be chosen such that the reconstructed signal level of $\hat{\mathbf{S}}^{(1)}_{\text{sig}}$ is approximately zero (i.e. at the noise level) outside its expected supports. Figure \ref{fig:fourier_all} and \ref{fig:best1} show the line integrals and the corresponding Fourier transforms of $(H,K)$-plane slices of the 4D signal. Specifically, we apply Radon transform at angle $\alpha = 0$ on the 4D signal summed over the $L$ and $\omega$ axes (i.e., $\mathcal{R}_{\alpha = 0}[\sum_{L,\omega} \bar{\mathbf{S}}^{\text{pred}}_{\text{expt}}(H,K,L,\omega)]$), yielding a 1D profile as a function of 
$K$. The corresponding Fourier transformation is obtained by taking the 2D FFT of $\sum_{L,\omega} \bar{\mathbf{S}}^{\text{pred}}_{\text{expt}}(H,K,L,\omega)$ and extracting the central horizontal slice $\xi_{H} = 0$ by the Fourier projection slice theorem. Line integrals are used for improved robustness to noise. The projection plots reveal a clear difference in intensity scales between the two reconstructed components and the effect of varying $\lambda$ on the energy distribution between the decomposed components. Our quantitative analysis confirms that $\lambda = 0.0005$ gives the best fitting in terms of various reconstruction statistics, see Table~\ref{tab:stats}. In contrast, it can be observed that a stronger constraint on the background intensity (e.g., $\lambda = 0.05$) causes an intensity leakage from the background to the signal (i.e., elevated signal intensity outside the expected support regions or voids in the background); this leakage effect is also evident in Figure \ref{fig:leakage}.

\subsubsection{Support of Kernel-Net $r$}
Another important consideration in our method is the choice of the width of the kernel represented by the second hyperparameter $r$, which sets the proper number of neighbors and controls the spatial support used by the learnable kernel and thus the convolved signal. It can be chosen using the following quantitative criteria.
\begin{itemize}
    \item \textbf{Physical consistency of the intensities}: The peak intensity across different spatial locations should align, with an experimentally observed retention of $80\%$, a commonly accepted threshold as a rule of thumb.  
    \item \textbf{Pixel-wise difference distribution}: Examination of the pixel value distributions and histograms of absolute differences between the original and reconstructed images. 
    \item \textbf{Statistical analysis}: The unfitted residual components can be tested using appropriate test statistics (e.g., the chi-squared test, or autocorrelation function).
\end{itemize}

Based on the three preceding quantitative criteria, we have explored $r$-values by fixing the optimal $\lambda = 0.0005$ and varying the support of the kernel network. As shown in Figure~\ref{fig:nr}, we plot the 45-degree slice of the reconstructed $4$D signal summed along the last two dimensions (i.e., $\sum_{L,\omega} \bar{\mathbf{S}}_{\text{expt}}^{\text{pred}}(H,K,L,\omega)$), for both the overall reconstructed signal and the separated components. Note that for $r = 4$, the four resolved peak intensities across different sample indices exhibit unphysical amplitude variations. In contrast, smaller values of $r$ yield peak patterns with physically consistent characteristics which infers a smaller $r$ is more favorable. We also show the learned kernels in Figure \ref{fig:learnt_kernels} for both cases. Moreover, the reconstruction accuracies for varying $r$ values at the fixed optimal $\lambda = 0.0005$ are also evident from the tight clustering around the line $y = x$ shown in Figure~\ref{fig:hist}. For a more detailed analysis of the differences between the reconstructed and ground truth signals across various energy levels, we refer readers to Figure \ref{fig:energy_vs_l.png} and \ref{fig:energy_comparison}. Furthermore, Table~\ref{tab:stats} summarizes various reconstruction statistics for different parameter configurations which points $r=2$ as the optimal choice. Collectively, these metrics consistently point to ${r = 2, \lambda = 0.0005}$ as the optimal choice for this data set.

\begin{table}[hbt!]
    \centering
    \begin{tabular}{rlllllll}
    \toprule
    $r$ & $\lambda$ & RMSE & PSNR & SSIM & RE & $\chi ^2$ & $\chi^2_{\text{pval}}$\\
    \midrule
    2 & 0.0005 & 4.122 & 37.998 & 0.911 & 0.118 & 6828.510 & 1.0 \\
    2 & 0.005 & 5.183 & 36.007 & 0.846 & 0.148 & 15315.516 & 0.0 \\
    3 & 0.05 & 5.265 & 35.870 & 0.806 & 0.151 & 16370.830 & 0.0 \\
    2 & 0.05 & 5.551 & 35.412 & 0.855 & 0.159 & 14785.543 & 0.0 \\
    4 & 0.05 & 6.523 & 34.010 & 0.767 & 0.187 & 20736.734 & 0.0 \\
    3 & 0.005 & 6.694 & 33.786 & 0.761 & 0.192 & 597478.826 & 0.0 \\
    3 & 0.0005 & 7.092 & 33.283 & 0.757 & 0.203 & 24967.078 & 0.0 \\
    4 & 0.0005 & 8.401 & 31.812 & 0.713 & 0.240 & 31636.260 & 0.0 \\
    4 & 0.005 & 9.630 & 30.627 & 0.691 & 0.276 & 38617.373 & 0.0 \\
    \bottomrule
    \end{tabular}
    \caption{\textbf{Summary statistics of reconstruction performance metrics for hyperparameter optimization using experimental La$_2$NiO$_4$ data}. Reported metrics include: Root Mean Square Error (RMSE), Peak Signal-to-Noise Ratio (PSNR), Structural Similarity Index Measure (SSIM), Relative Error (RE), Chi-square statistic, and corresponding p-values. All metrics consistently points to $\{r=2,\lambda = 0.0005\}$ as the optimal choice.}
    \label{tab:stats}
\end{table}

These results underscore the critical need for domain-specific hyperparameter optimization to ensure consistent and reliable separation outcomes.

\section{Discussion and Future Directions}\label{Sec:4}


Through comprehensive benchmarking on both synthetic and experimental data, we show that the proposed method effectively captures complex peak profiles even in the presence of strong background signals across the full four-dimensional space. 
While this work relies on simulated signals $\mathbf{S}_{\text{sim}}$, future studies may explore replacing them with surrogate peak models, for example, using peak-finding algorithms with simple Gaussian kernels, to eliminate the need for theoretically computed spectra. The proposed framework can also be generalized to a multi-INR architecture capable of handling more than two source contributions.
\\

Our proposed architecture also provides an effective denoising and data compression method. The denoising effect arises from the inherent low-frequency bias of INRs, which naturally suppress noise as a byproduct. For compression performance, our original non-sparse inelastic neutron scattering dataset of La$_2$NiO$_4$, which totaled $\sim$1.19GB in pixel representations, could be efficiently represented and separated using significantly smaller neural networks requiring just 1.503MB of total storage for the optimal configuration. This is a compression ratio of 792:1, corresponding to a $99.87\%$ reduction in storage requirements. Table~\ref{tab:model_size} compares the original data size and the compact model sizes across various parameter configurations. Our method thus holds promise for dramatic compression of experimental data while preserving a controllable trade-off between compression and accuracy. By providing lightweight data snapshots, it can assist real-time decision making to help guide adaptive measurement strategies in time-constrained experimental settings. This will be explored further in future work.
\\

While the method is highly useful for signal extraction, denoising and compression, it is worth pointing out that different scientific goals demand varying levels of fidelity in separating signal and background components. The motivations for separating data sources can be manifold. For instance, it might be of interest to extract a signal corresponding to a particular excitation with high precision. In this case, even though modeling of the background is necessary, it need not be a priority. In another scenario, where the dominant spectral feature is well described by simulations, signatures of other less understood excitations buried in the background could be of greater interest, and the background would then require more accurate modeling. As an example, although our focus in this study has been on single-magnon excitations, multi-magnon excitations that contribute to the continuum background are also of scientific interest. In short, we should keep in mind that the unmodeled components of a dataset can carry valuable information.
\\

Finally, we note that the presented work is foremost a proof of concept. It still entails a large set of degrees of freedom to optimize the framework for a given scientific problem. Even in the current example in the modeling of single-magnon excitations in La$_2$NiO$_4$, there is a wide range of sensible refinements available. An important element for optimization without largely changing the framework fundamentally is the loss function, given in Eq.~\ref{eq:loss_fct}. Modifying the loss function based on the scientific problem to be modeled will improve further the success of this framework. Such modifications may include for example functional transformations of the experimental and simulated signals, $\mathcal{L} = \| f(\mathbf{S}_{\text{expt}}^{\ast}) - f(\bar{\mathbf{S}}_{\text{expt}}^{\text{pred}}) \|_2 + \lambda \|f(\mathbf{B}_{\hat{\theta}})\|_2$ , e.g. to enhance the weighting of weaker signals, or multiplicative weights $w$ varying throughout the parameter space such that $\mathcal{L} = \| w_{\mathbf{Q},\omega}\,\mathbf{S}_{\text{expt}}^{\ast} - w_{\mathbf{Q},\omega}\,\bar{\mathbf{S}}_{\text{expt}}^{\text{pred}} \|_2 + \lambda \|w_{\mathbf{Q},\omega}\,\mathbf{B}_{\hat{\theta}}\|_2$. For example, for the presented experimental data $\mathbf{S}_{\text{expt}}$, previous work \cite{chitturi2023capturing} found that modeling $\mathrm{log}(\mathbf{S}_{\text{expt}}+1)$ rather than $\mathbf{S}_{\text{expt}}$ yielded a significantly more stable convergence and with results closer to the ground truth. Alternatively, the parameters $J$ and $J_p$ mostly affect the parameter space with larger $\omega$ where the signals however are not as strong. This implies that regions with low $\omega$ where the signals are strongest, are more heavily weighted even though they contain less information about the parametrization of this particular scientific model. Hence, using a weight function which increases with $\omega$ may also lead to improvements for modeling the physics in this system.

\section{Conclusion}
We have shown how the fundamental problem of source separation from multiple source contributions can be approached from a fresh perspective by training a dual implicit neural representation. We demonstrate the utility and efficacy of our framework by applying it to a 4D inelastic neutron spectroscopy dataset, supplemented with simulations. In this way, we successfully extract  the single-magnon excitations from the inelastic neutron scattering data of a Mott insulator. This work provides a powerful and versatile framework for disentangling signals from multiple sources while simultaneously providing denoising and compression capabilities, contributing to the advancement of machine learning methods for scientific data analysis and source separation.

\section*{Acknowledgments}
This work was primarily supported by the U.S. Department of
Energy, Office of Science, Basic Energy Sciences under Award No.
DE-SC0022216.
Y.N., Z.C., and J.J.T. were partially supported by the Department of Energy, Laboratory Directed Research and Development program at SLAC National Accelerator Laboratory, under contract DE-AC02-76SF00515. C.P. was supported by the Department of Energy, Office of Science, Basic Energy Sciences, Materials Sciences, and Engineering Division under Contract No.DE-AC02-76SF00515. This research used resources at the Spallation Neutron Source, a DOE Office of Science User Facility operated by the Oak Ridge National Laboratory. This research used resources of the National Energy Research Scientific Computing Center, a DOE Office of Science User Facility supported by the Office of Science of the U.S. Department of Energy under Contract DE-AC02-05CH11231 using NERSC award BES-ERCAP0026843. The Linac Coherent Light Source (LCLS), SLAC National Accelerator Laboratory, is supported by the U.S. Department of Energy, Office of Science, Office of Basic Energy Sciences under Contract No. DE-AC02-76SF00515.

During the preparation of this work, the authors used large language model ChatGPT by OpenAI in order to refine the language and enhance the readability of this paper. After using this tool or service, the authors reviewed and edited the content as needed and take full responsibility for the content of the publication.

\bibliographystyle{abbrv}
\bibliography{ref}  

\clearpage
\section*{Appendix A}
\renewcommand{\thefigure}{A\arabic{figure}}
\setcounter{figure}{0}
\renewcommand{\theequation}{A\arabic{equation}}
\setcounter{equation}{0}
\renewcommand{\thetable}{A\arabic{table}}
\setcounter{table}{0}
\begin{figure}[hbt!]
    \centering
    \includegraphics[width=\linewidth]{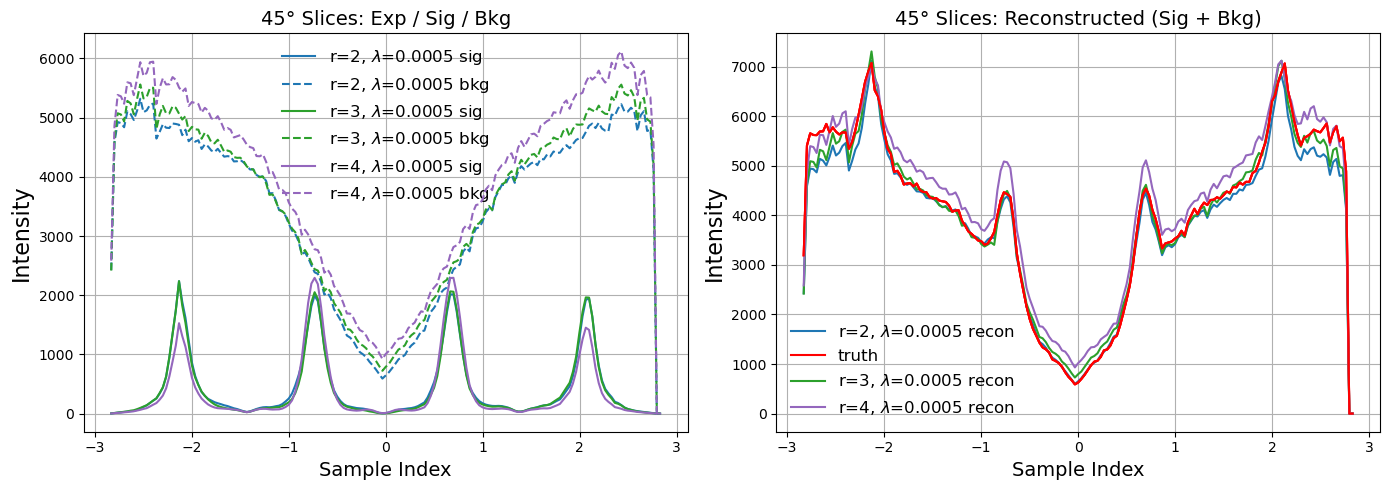}
    \caption{\textbf{Diagonal slices of 2D reconstructions with varying $r$.} We plot the 1D slice of the reconstructions by taking the diagonal (45-degree) slice across the reconstructed signals summed along the last two dimensions $\sum\limits_{L,\omega}\bar{\mathbf{S}}^{\text{pred}}_{\text{expt}}(H,K,L,\omega),\sum\limits_{L,\omega}\hat{\mathbf{S}}^{(1)}_{\text{sig}}(H,K,L,\omega)$ and $\sum\limits_{L,\omega}\hat{\mathbf{S}}^{(2)}_{\text{sig}}(H,K,L,\omega)$ respectively. The left panel shows the decomposed signal and background components separately, while the right panel displays the combined reconstructions. Using the optimal $\lambda = 0.0005$, we compare results using varying $r \in \{2,3,4\}$. Note that for $r=4$, the four resolved peak intensities across the various sample indices exhibit unphysical variations in amplitudes.}
    \label{fig:nr}
\end{figure}

\begin{figure}[hbt!]
    \centering
    \includegraphics[width=\linewidth]{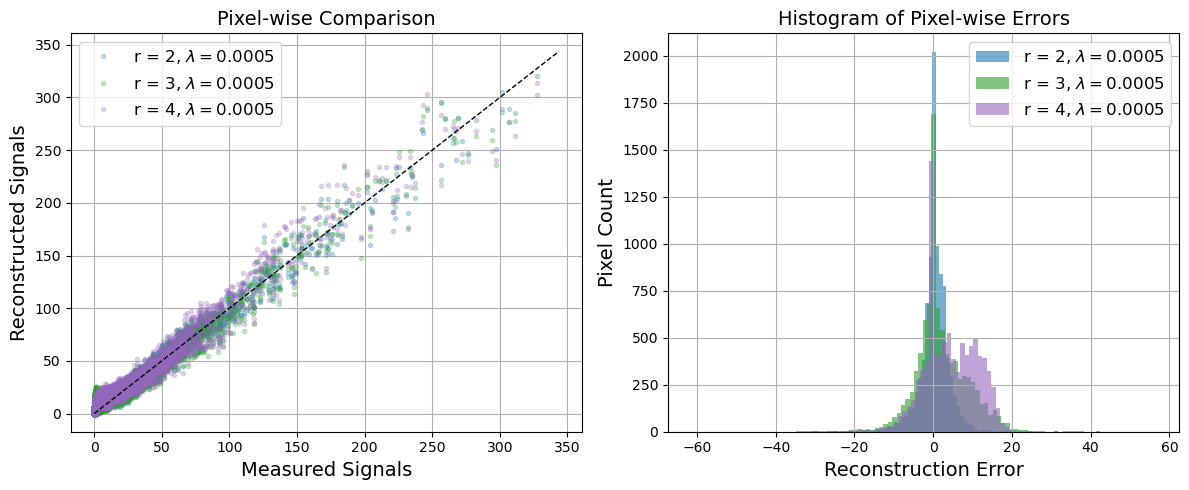}
    \caption{\textbf{Pixel-wise differences.} For a fixed $\lambda = 0.0005$, we analyze the pixel wise differences resulting from using difference choices of $r$. (a) shows the scatter plot where the flattened pixel value of the measurement is on the x-axis while the flattened reconstructed pixel value is on the y-axis. A perfect fit would yield the line $y=x$. In panel (b), we display the histogram of pixel-wise errors (differences between the original and reconstructed values) for each neighbor-range setting. In both analysis, $r=4$ is not favorable, suggesting overfitting to error.}
    \label{fig:hist}
\end{figure}

\begin{figure}[hbt!]
    \centering
    \includegraphics[width=\textwidth]{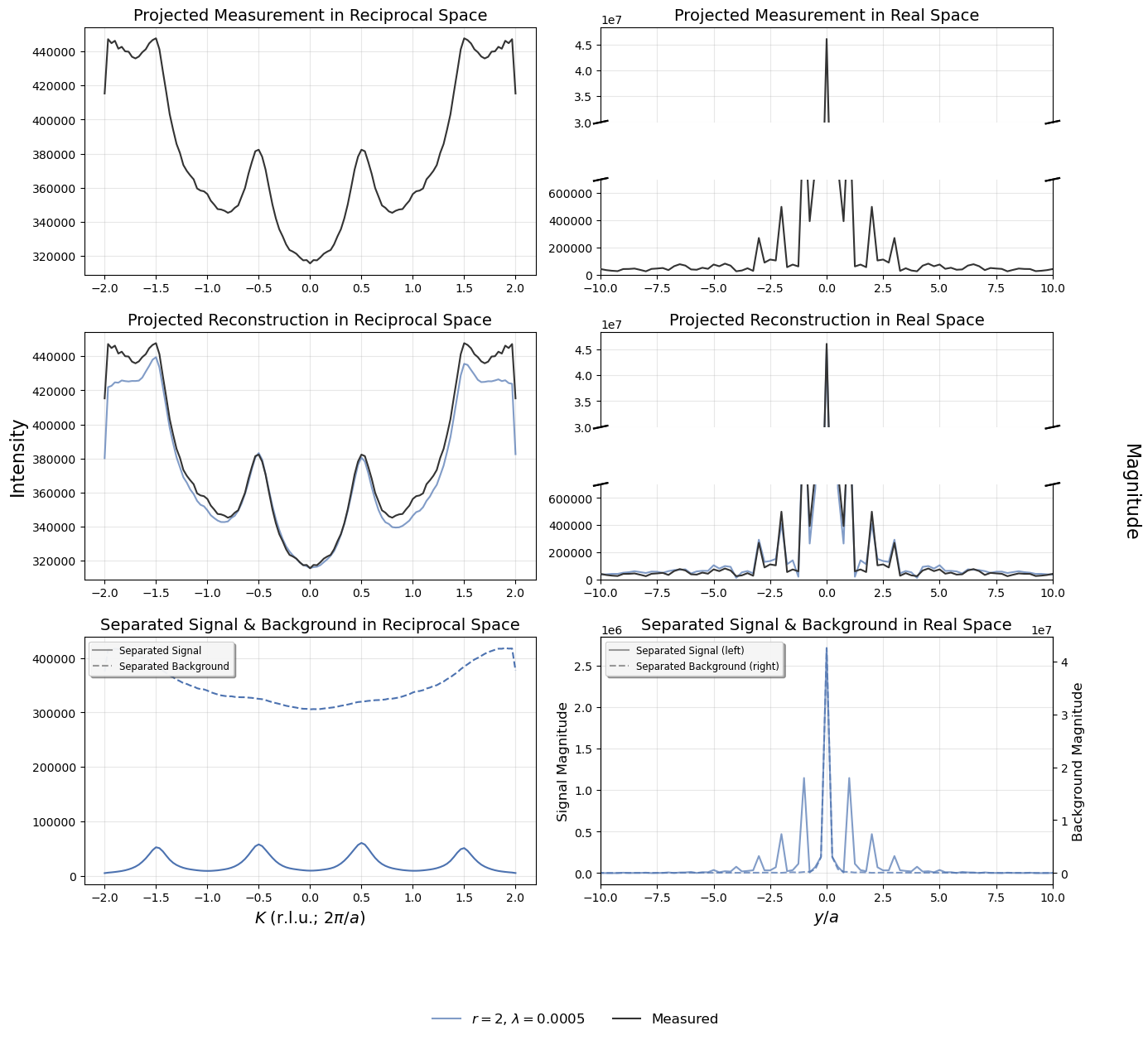}
    \caption{Central slice of the reconstructed signal and background components using the optimal configuration ($r = 2$, $\lambda = 0.0005$), shown alongside their corresponding magnitude in both reciprocal space $\mathcal{R}_{\alpha = 0}(\sum_{L,\omega}\bar{\mathbf{S}}^{\text{pred}}_{\text{expt}}(H,K,L,\omega))$ (left column) and its Fourier transformation $\mathcal{F}[\sum_{L,\omega}\bar{\mathbf{S}}^{\text{pred}}_{\text{expt}}(H,K,L,\omega)](\xi_H=0)$ (right column). The first row displays the central slice of the experimental measurements, the second row shows that of the total reconstruction, and the third row presents the separated signal and background components. This hyperparameter setting achieves the best empirical separation of signal and background.}
    \label{fig:best1}
\end{figure}

\begin{figure}[hbt!]
	\centering

     \includegraphics[width = 0.9\linewidth]{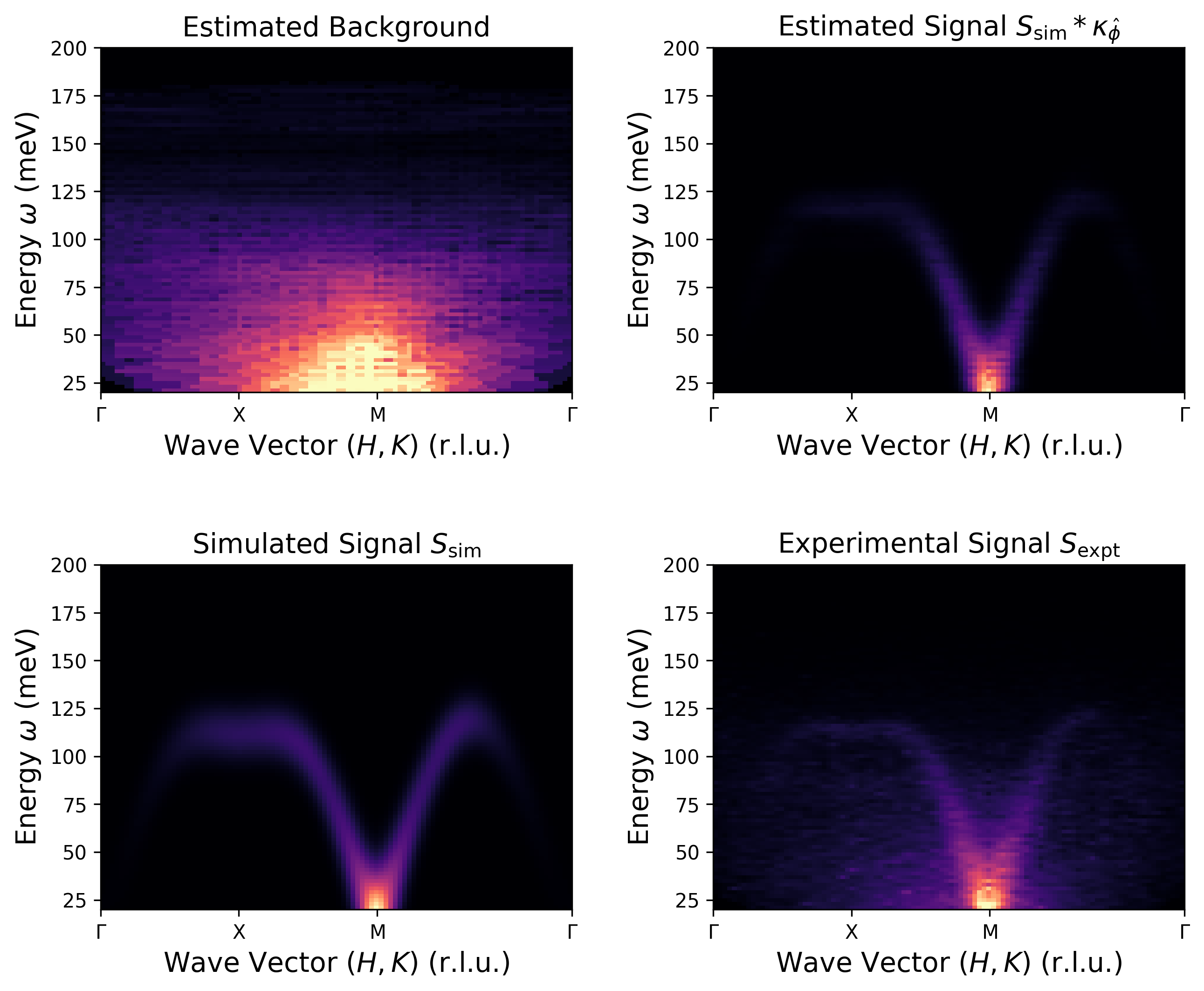}
	\caption{
    Spatial lattice structure of the measured material, La$_2$NiO$_4$ (lower right) and the simulated signal (lower left), shown alongside the extracted background (upper left) and the extracted estimated signal (upper right).
    }
	\label{fig:bestrecon}
\end{figure}

\begin{figure}[hbt!]
    \centering
    \begin{subfigure}[b]{0.8\linewidth}
        \centering
        \includegraphics[width=\linewidth]{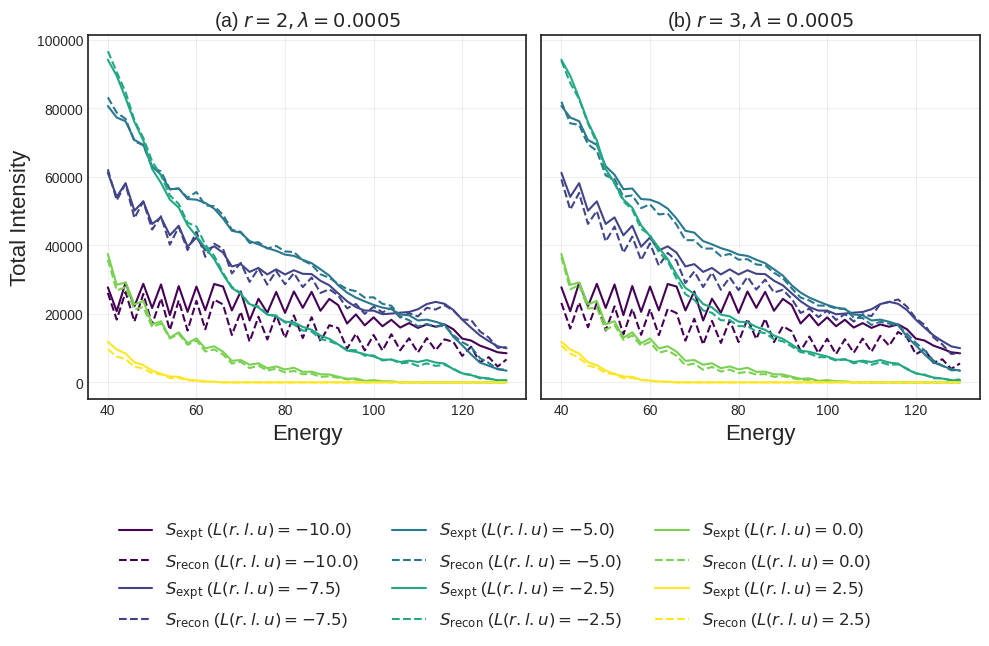}
        \caption*{(a) \text{\text{$r=2, \lambda = 0.0005$}}}
    \end{subfigure}
    \hfill
    \begin{subfigure}[b]{0.8\linewidth}
        \centering
        \includegraphics[width=\linewidth]{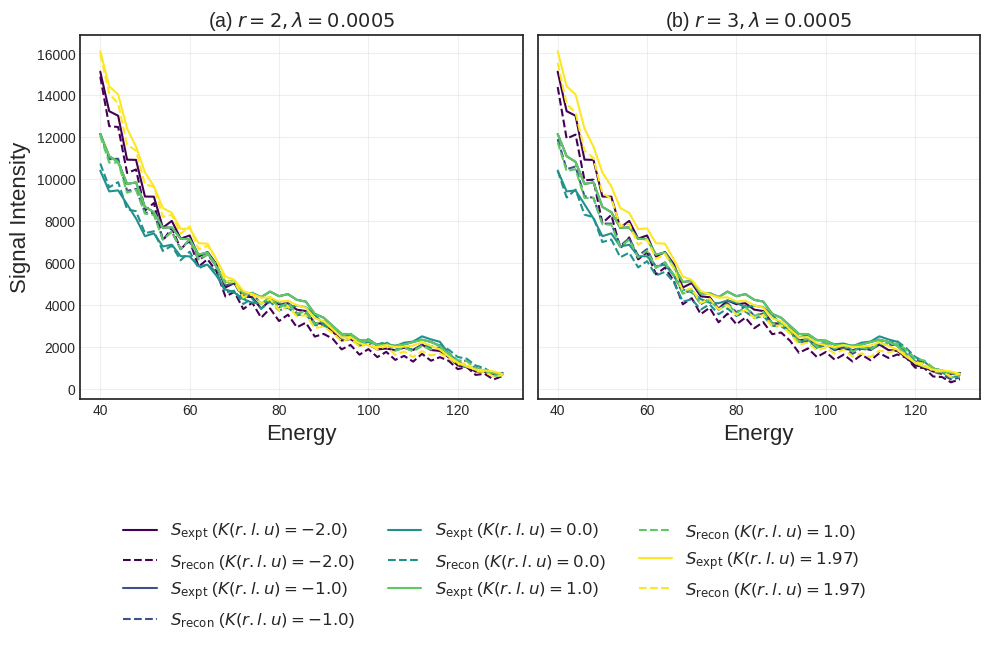}
        \caption*{(b) \text{$r=3, \lambda = 0.0005$}}
    \end{subfigure}
    \caption{Differences between the reconstructed and ground-truth signals across energy. (a) Intensities summed over $(H,K)$ as a function of energy, with each curve corresponding to a different $L$ index. (b)Intensities summed over $(H,L)$, with each curve corresponding to a different $K$. The $x$-axis represents energy $\omega$, and the $y$-axis shows total intensity for each of the measurements $\textbf{S}^*_{\text{expt}}$ (solid lines) and reconstruction $\bar{\textbf{S}}^{\text{pred}}_{\text{expt}} = \hat{\textbf{S}}^{(1)}_{\text{sig}}+\hat{\textbf{S}}^{(2)}_{\text{sig}}$ (dashed lines).} 
    \label{fig:energy_vs_l.png}
\end{figure}

\begin{figure}[htb!]
    \centering

    \begin{subfigure}[b]{\linewidth}
        \centering
        \includegraphics[width=\linewidth]{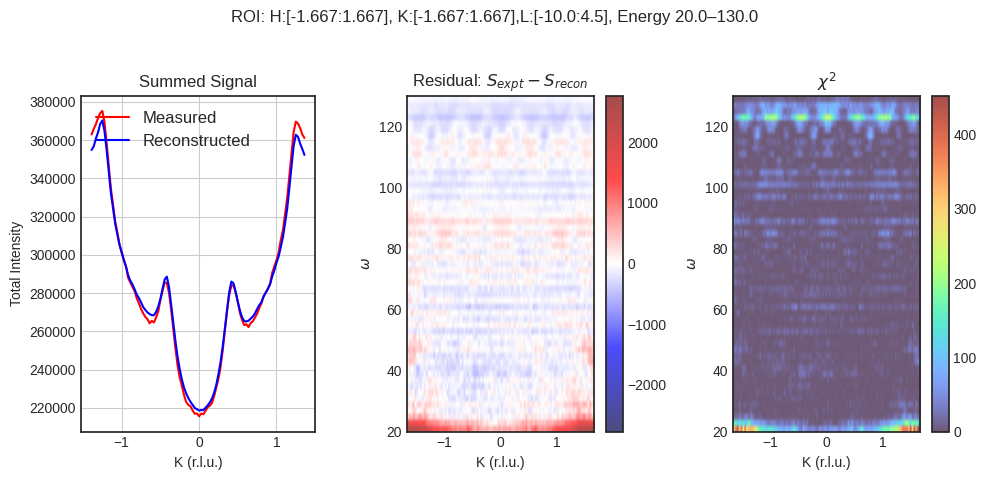}
        \caption*{(a) \text{\text{$r=2, \lambda = 0.0005$}}}
    \end{subfigure}
    \hfill
    \begin{subfigure}[b]{\linewidth}
        \centering
        \includegraphics[width=\linewidth]{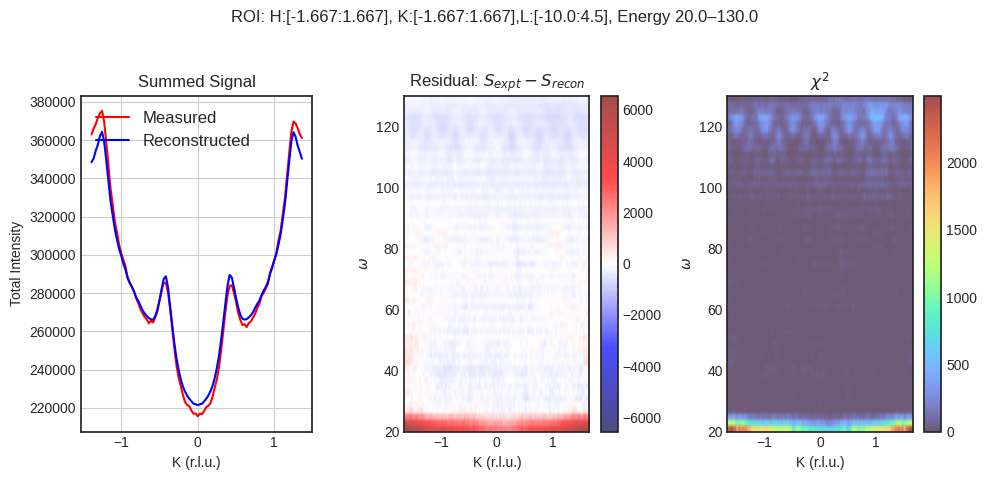}
        \caption*{(b) \text{$r=3, \lambda = 0.0005$}}
    \end{subfigure}
    \caption{Reconstruction errors for energies in the range $[20, 130]$ meV, summed along $(H, L)$ as a function of $K$ and $\omega$. The $x$-axis represents $K$, and the $y$-axis represents energy $\omega$. To avoid boundary artifacts, only the central region of the data is shown (i.e., $H,K \in [-1.667, 1.667]$), excluding the outer edges of the original range $H,K \in [-2, 2]$. Panels (a) and (b) show results for different choices of $r$ and $\lambda$. Each set of three panels displays the reconstructed intensities, the measured intensities, the residuals across $K$ and $\omega$, and the corresponding $\chi^2$ statistics.}
    \label{fig:energy_comparison}
\end{figure}

\begin{figure}[hbt!]
    \centering
        \begin{subfigure}[a]{\linewidth}
        \includegraphics[width=\linewidth]{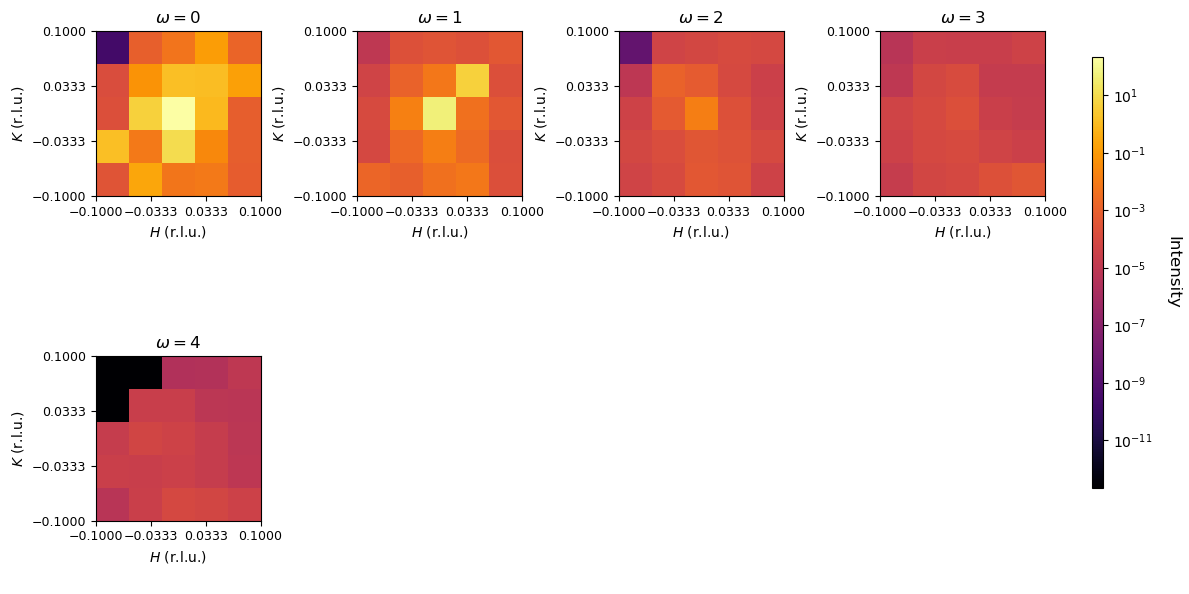}
       \caption{}
        \label{fig:kernel_nr2}
    \end{subfigure}

    \vspace{0.5em} 

    \begin{subfigure}[b]{\linewidth}
        \includegraphics[width=\linewidth]{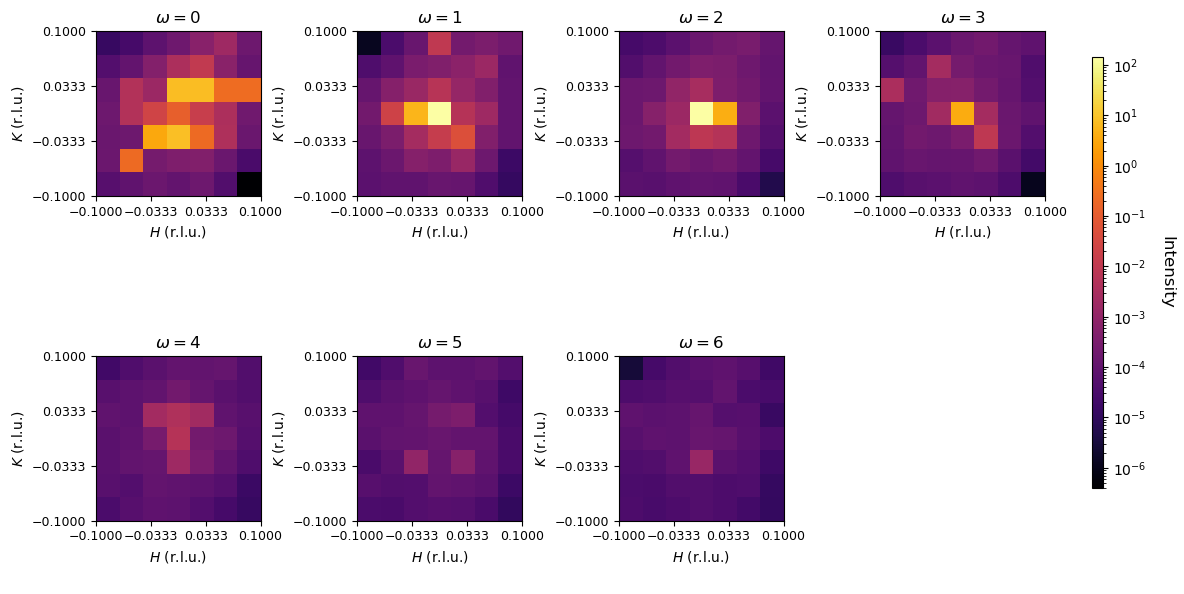}
        \label{fig:kernel_nr3}
        \caption{}
    \end{subfigure}
    \caption{Visualization of the learned convolutional kernels averaged over all $\mathbf{Q}$ and $\omega$. For $r=2$, the effective kernel has a grid size of $5\times5\times5\times5$ at each $\mathbf{Q}$ and $\omega$; for $r=3$, the grid size is $7\times7\times7\times7$. Panels (a) and (b) show $\omega$ slices summed over $L$ for $r=2$ and $r=3$, respectively. Note that for $r=3$, the effective kernel support exceeds the representational capacity of $r=2$’s smaller window, suggesting that $r=3$ can capture spatially extended distortions more effectively when the signal of interest exhibit such larger scale distortions. However, $r = 2$ is sufficient for this dataset, as most entries of the $r = 3$ kernel near the boundaries are close to zero.}
    \label{fig:learnt_kernels}
\end{figure}

\clearpage
\section*{Appendix B}\label{Appendix B}
\setcounter{figure}{0}
\setcounter{equation}{0}
\setcounter{table}{0}
\setcounter{section}{0}
\renewcommand{\thetable}{B\arabic{table}}
\renewcommand{\thefigure}{B\arabic{figure}}
\renewcommand{\theequation}{B\arabic{equation}}
\renewcommand{\thesection}{B\arabic{section}}
\renewcommand{\thesubsection}{B\arabic{subsection}}
We provide a summary of our proposed dual-INR training procedure in Algorithm 1, which outlines the main steps involved in the optimization pipeline. 

\begin{algorithm}[H]
\SetAlgoLined
\KwIn{Pre-trained \texttt{SpecNeuralRepr}, center points $\mathbf{x}_c$, neighbor points $\mathbf{x}_n$, experimental data $\mathbf{S}_{\text{expt}}^*$, regularization weight $\lambda$, window size of the kernel $r$.}
\KwOut{Total training loss}

\tcc{Forward Pass}
$\boldsymbol{\kappa}_{r,\hat{\phi}} \gets \texttt{KernelNet}(\mathbf{x}_c;r) $ \tcp*{Learned kernel weights at center points}
$\mathbf{S}_{\text{sim}} \gets \texttt{SpecNeuralRepr}(\mathbf{x}_n)$ \tcp*{Simulated signal at neighbor points}
$\hat{\mathbf{S}}_{\text{sig}}^{(1)} \gets \texttt{einsum}(\boldsymbol{\kappa}, \mathbf{S}_{\text{sim}})$ \tcp*{Masked kernel-signal convolution}
$\hat{\mathbf{S}}_{\text{sig}}^{(2)} \gets \texttt{BkgdNet}(\mathbf{x}_c)$ \tcp*{Learned background signal}
$\bar{\mathbf{S}}_{\text{expt}}^{\text{pred}} \gets \hat{\mathbf{S}}_{\text{sig}}^{(1)} + \hat{\mathbf{S}}_{\text{sig}}^{(2)}$ \tcp*{Final reconstruction}

\tcc{Loss Computation}
$\text{loss}_{\text{reconst}} \gets \texttt{MSE}(\bar{\mathbf{S}}_{\text{expt}}^{\text{pred}}, \mathbf{S}_{\text{expt}}^*)$ \tcp*{Reconstruction error}
$\text{loss}_{\text{bkg}} \gets \lambda \cdot \texttt{Mean}\left((\hat{\mathbf{S}}_{\text{sig}}^{(2)})^2\right)$ \tcp*{Background magnitude penalty}
$\text{loss} \gets \text{loss}_{\text{reconst}} + \text{loss}_{\text{bkg}}$ \tcp*{Total training loss}

\Return $\text{loss}$
\caption{Training step for the proposed dual INR for image source separation.}
\end{algorithm}

This includes calls to multiple INRs used to model the simulated signal, the kernel, and the background components. Table B.1 presents the detailed architectures of the neural networks used for each INR, including layer configurations and activation functions.

\begin{table}[hb!]
\centering
\begin{tabular}{>{\bfseries}l m{9cm}}
\toprule
Component & Structure Summary \\
\midrule
KernelNet &
\begin{itemize}
  \item Input: $(\mathbf{Q},\omega) \in \mathbb{R}^4$
  \item Subnet: SirenNet with $L$ layers, ReLU final activation
  \item Followed by: Linear $\rightarrow$ ReLU $\rightarrow$ Linear $\rightarrow$ Softmax 
  \item Output: kernel weights $\boldsymbol{\kappa}(\mathbf{Q},\omega) \in \mathbb{R}$, where masking is applied based on the hyperparameter $r$, which defines the window size of the kernel.
\end{itemize}
\\
\addlinespace
BkgdNet &
\begin{itemize}
  \item Input: $(\mathbf{Q},\omega) \in \mathbb{R}^4$
  \item Single hidden layer SirenNet with ReLU as final activation
  \item Output: background signal $\hat{\mathbf{S}}^{(2)}_{\text{sig}}(\mathbf{Q},\omega) $
\end{itemize}
\\
\addlinespace
SirenNet &
\begin{itemize}
  \item Sinusoidal Representation Network (SIREN \cite{SIREN})
  \item First layer initialized with frequency scaling $w_0=30$
  \item Each layer: Linear $\rightarrow$ Sine activation
  \item Optional final activation (e.g., ReLU)
\end{itemize}

\\
\bottomrule
\end{tabular}
\caption{Summary of neural network components in the dual INR training pipeline.}
\label{table:NN summary}
\end{table}

\subsection{Model size comparison}

The proposed dual INR framework also serves as an efficient lossy compression method. The weights of the trained INRs encode information about the raw signal. As a result, instead of storing the full raw data, we can achieve compression by storing only the INR weights. To support this idea, we include a comparison of the INR model sizes used in our numerical experiments and the original data sizes under various hyperparameter settings.

\begin{table}[hbt!]
  \caption{Model size comparison for using different neighbor ranges $r$. The original data size in pixel representations is $\sim$ 1.19G, the data is non-sparse.}

  \centering
  \begin{minipage}{0.45\linewidth}
    \centering
    \begin{tabular}{lll}
      \multicolumn{3}{c}{$r=2$} \\
      \toprule
      Name & Type & Size (byte) \\
      \midrule
      kernel\_net    & KernelNet       & 841k \\
      bkgd\_net      & SirenNet        & 133k \\
      forward\_model & SpecNeuralRepr  & 529k \\
      \bottomrule
    \end{tabular}
    \label{tab:2neighb}
  \end{minipage}
  
  \vspace{2em}
  
  \begin{minipage}{0.45\linewidth}
    \centering
    \begin{tabular}{lll}
      \multicolumn{3}{c}{$r=3$} \\
      \toprule
      Name & Type & Size (byte) \\
      \midrule
      kernel\_net    & KernelNet       & 1.8M \\
      bkgd\_net      & SirenNet        & 133k \\
      forward\_model & SpecNeuralRepr  & 529k \\
      \bottomrule
    \end{tabular}
    \label{tab:3neighb}
  \end{minipage}

  \vspace{2em}

  \begin{minipage}{0.45\linewidth}
    \centering
    \begin{tabular}{lll}
      \multicolumn{3}{c}{$r=4$} \\
      \toprule
      Name & Type & Size (byte) \\
      \midrule
      kernel\_net    & KernelNet       & 3.9M \\
      bkgd\_net      & SirenNet        & 133k \\
      forward\_model & SpecNeuralRepr  & 529k \\
      \bottomrule
    \end{tabular}
    \label{tab:4neighb}
  \end{minipage}
  \label{tab:model_size}
\end{table}


\end{document}